\documentclass[runningheads]{llncs}

\usepackage{eccv}

\usepackage{eccvabbrv}

\usepackage{graphicx}
\usepackage{booktabs}
\usepackage[table]{xcolor}
\usepackage{colortbl}        
\usepackage{pifont}          
\newcommand{\cmark}{\ding{51}}
\newcommand{\xmark}{\ding{55}}
\usepackage{algorithm}
\usepackage{algorithmicx}
\usepackage{algpseudocode}
\usepackage{xcolor}
\usepackage{multirow}
\usepackage{enumitem}
\usepackage[accsupp]{axessibility}  

\usepackage[pagebackref,breaklinks,colorlinks,citecolor=eccvblue]{hyperref}

\usepackage{orcidlink}

\begin{document}

\title{\textbf{\textit{1.x-Distill}}: Breaking the Diversity, Quality, and Efficiency Barrier in Distribution Matching Distillation} 




\titlerunning{1.x-Distill}

\author{
  Haoyu Li\inst{1}\thanks{Equal Contribution} \and
  Tingyan Wen\inst{1}\protect\footnotemark[1] \and 
  Lin Qi\inst{2}\thanks{Corresponding Author} \and
  Zhe Wu\inst{2} \and
  Yihuang Chen\inst{2} \and
  Xing Zhou\inst{2} \and
  Lifei Zhu\inst{2} \and
  XueQian Wang\inst{1} \and
  Kai Zhang\inst{1}\protect\footnotemark[2]
}
\authorrunning{H. Li et al.}

\institute{
  Tsinghua University \and
  Central Media Technology Institute, Huawei 
}

\maketitle

\begin{center}
  \centering
  {\tt\small \url{https://thu-accdiff.github.io/1.x-distill-page/}}
\end{center}

\begin{abstract}
\label{sec:abstract}
Diffusion models produce high-quality text-to-image results, but their iterative denoising is computationally expensive.
Distribution Matching Distillation (DMD) emerges as a promising path to few-step distillation, but suffers from diversity collapse and fidelity degradation when reduced to two steps or fewer. We present \textbf{\textit{1.x-Distill}}, the first fractional-step distillation framework that breaks the integer-step constraint of prior few-step methods and establishes 1.x-step generation as a practical regime for distilled diffusion models.
Specifically, we first analyze the overlooked role of teacher CFG in DMD and introduce a simple yet effective modification to suppress mode collapse. Then, to improve performance under extreme steps, we introduce \emph{Stagewise Focused Distillation}, a two-stage strategy that learns coarse structure through diversity-preserving distribution matching and refines details with inference-consistent adversarial distillation. Furthermore, we design a lightweight compensation module for \emph{Distill--Cache co-Training}, which naturally incorporates block-level caching into our distillation pipeline.
Experiments on SD3-Medium and SD3.5-Large show that \textbf{\textit{1.x-Distill}} surpasses prior few-step methods, achieving better quality and diversity at 1.67 and 1.74 effective NFEs, respectively, with up to $\mathbf{33\times}$ speedup over original 28×2 NFE sampling.
\end{abstract}
\section{Introduction} \label{sec:intro}
\begin{figure}[t]
  \centering
    \includegraphics[width=\columnwidth]{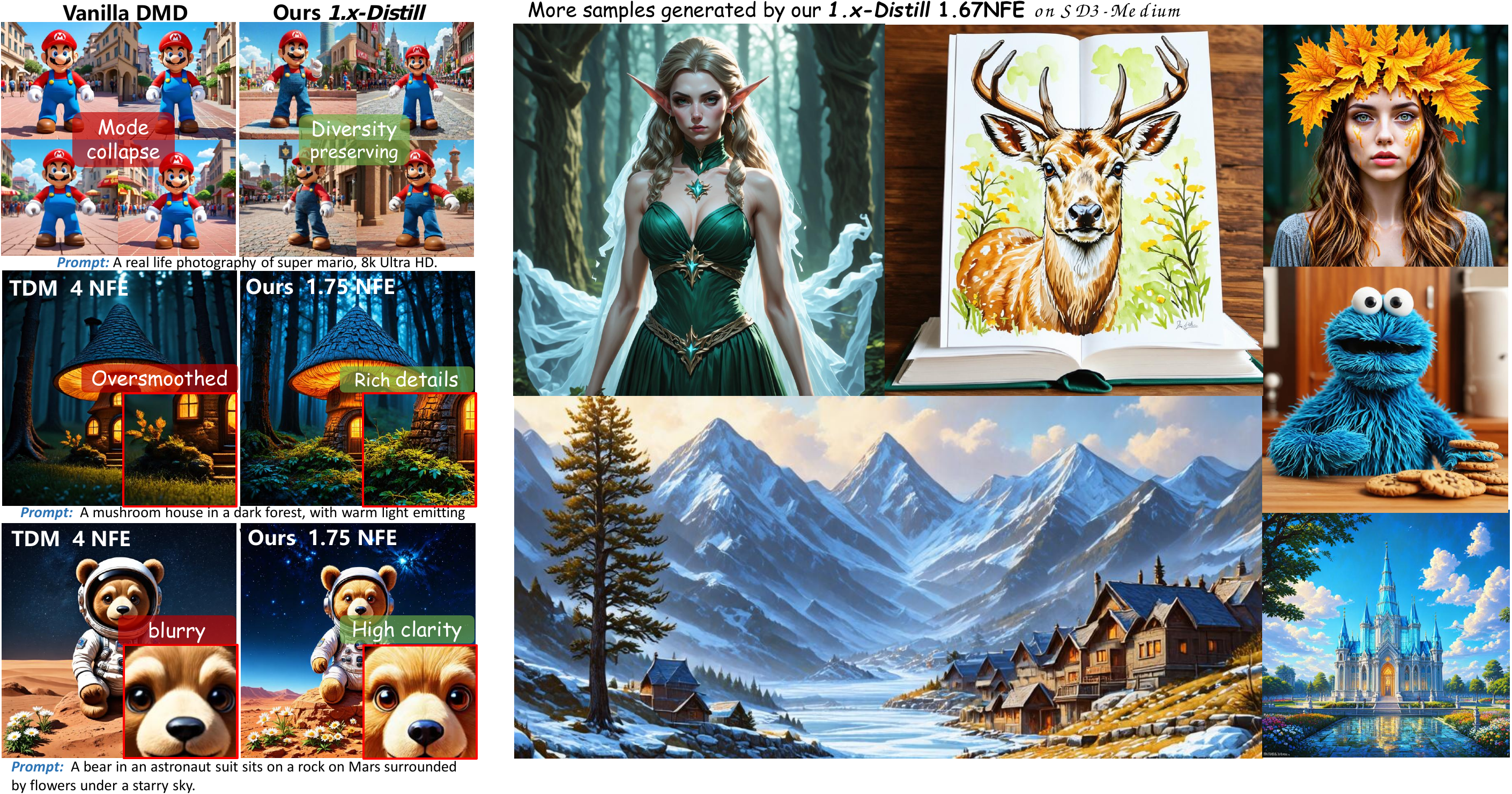} 
  \caption{\textbf{Visual results.} \textbf{\textit{1.x-Distill}} mitigates the mode collapse and quality degradation of vanilla DMD under extreme step reduction, delivering superior few-step results.}
  \label{fig:introduction}
\end{figure}

Diffusion models\cite{dm-ddpm,dm-beat-gan,sd,sdxl,sd3} have become the dominant paradigm at high-resolution image generation, but their iterative sampling steps leads to high computational cost. To mitigate this issue, recent research has actively explored step distillation\cite{progressive,add,ladd,dmd,dmd2,hypersd,cm,meanflow}, which distill a multi-step pretrained diffusion model into a few-step generator. Among them, Distribution matching distillation (DMD)\cite{dmd,dmd2} reduces the student’s sampling to a few steps by matching the output distribution of the teacher, and have demonstrated strong effectiveness on large-scale models.

However, as shown in the left panel of Fig.~\ref{fig:introduction}, existing distribution matching methods~\cite{dmd2,tdm,senseflow} face two major bottlenecks when pushed to two-step or fewer sampling. (1) Compared to trajectory-based distillation\cite{lcm,rcm,meanflow}, DMD series suffer from severe diversity degradation. (2) Extreme step reduction forces each denoising step to carry more semantic and visual responsibility, which leads to pronounced quality degradation in the generated images. 

While \emph{mode collapse} in DMD is often attributed to the reverse KL formulation\cite{dmdx,rcm}, we provide a complementary perspective by analyzing the role of Classifier-Free Guidance (CFG)\cite{cfg} during training. 
We observe that the strong CFG used in the real score prediction at high-noise timesteps can prematurely bias the student toward dominant modes. 
Rather than previous methods\cite{dmdr,rcm} introducing additional training efforts to explicitly encourage mode-covering, we control the teacher guidance in a timestep-aware manner within the DMD framework. This simple yet effective modification improves the student diversity without extra modules or supervision.

To further overcome the quality bottleneck, we propose \textbf{Stagewise Focused Distillation (SFD)}. Student optimization is inherently stage-dependent, shifting from global structure formation to fine-detail refinement. Prior methods\cite{dmd2,senseflow} typically use uniform objectives throughout distillation, overlooking this training dynamics and leading to poor-quality generation. We therefore argue that a strong student should learn stage specific skills, and design SFD to align training objectives.
In the early stage, we apply non-uniform importance sampling and control the guidance in distribution matching to build structural stability and diversity. In the later stage, we switch to pixel-space adversarial distillation to enhance fine details. 
Distinct from prior approaches\cite{ladd,sd3.5flash,senseflow}, our adversarial distillation is formulated in a training-inference consistent manner to refine generation without disrupting the structure. As a result, SFD makes two step sampling both structurally reliable and detail rich. 

Even with high-quality 2-step sampling, further acceleration is still limited by heavy block-level computation. Since adjacent denoising steps are often similar, recomputing all blocks at every step is largely redundant, making cross-step reuse a natural complementary direction. However, existing cache methods\cite{teacache,taylorseers,delta-dit} are mostly tailored to standard multi-step diffusion, and directly applying them to few-step distilled models causes visual degradation due to large reuse error.

To address this, we propose \textbf{Distill--Cache co-Training (DCT)}, the first approach to integrate block-level caching into few-step distillation through joint reuse and error correction. Notably, the second stage of SFD naturally provides recovery training for cache accelerated inference on the final step, making fractional step sampling feasible without extra complexity.

In summary, our contributions are as follows:
\begin{itemize}
\item We revisit the overlooked role of teacher CFG in DMD and introduce a simple yet effective modification to preserve sampling diversity.
\item We propose \textbf{\textit{1.x-Distill}}, the first distillation framework that breaks the conventional integer-step constraint and achieves diverse, high-quality 1.x-step image generation.
\item We introduce two key techniques. \textbf{SFD} aligns training objectives with stage-dependent learning dynamics to improve extreme few-step quality, while \textbf{DCT} integrates block-level caching with reuse-error correction 
to eliminate redundant computation.

\item We achieve SOTA few-step performance on \emph{SD3-Medium} and \emph{SD3.5-Large}, attaining strong image quality with improved diversity at 1.67 and 1.74 effective NFE respectively, and up to $33\times$ speedup over 28×2 NFE sampling.

\end{itemize}

\section{Related Work}

\subsection{Few-Step Diffusion Distillation}
Existing Few-step distillation methods can be broadly categorized into trajectory-based and distribution-based approaches.
\textbf{Trajectory-based} methods aim to train a student to reproduce the PF-ODE trajectory of a teacher
model. Early works such as \emph{Progressive Distillation}\cite{progressive,sdxl-lightning} reduce the number of sampling steps in a staged manner but suffer from high training cost and accumulated error. Another representative line, \emph{Consistency Distillation}\cite{cm,lcm,meanflow} enforces self-consistency along the trajectory. These methods require careful formulations and non-trivial implementation on large-scale models.
\textbf{Distribution-based} methods aim to train a few-step student by aligning its output distribution with the target distribution. \emph{Adversarial Distillation}\cite{sdxl-lightning,add,ladd} can be viewed as a distribution-based approach, which introduces GAN-based\cite{gan} objectives to diffusion distillation. Another promising direction explores \emph{score distillation}\cite{prolificdreamer,diff-instruct, dmd}. Representative method DMD\cite{dmd} aligns the student distribution with the teacher via a reverse-KL objective and has become a strong baseline for large-scale few-step generation. Recent works such as DMD2\cite{dmd2}, DMDX\cite{dmdx}, TDM\cite{tdm}, SenseFlow\cite{senseflow} and Decoupled-DMD\cite{decoupleddmd} further improve DMD performance by enhancing training within original framework or combining additional objectives. Nevertheless, these methods still suffer from noticeable quality degradation under extreme step budgets for high-resolution generation.

\subsection{Cache Accelerator for Diffusion Models}
Cache-based acceleration has emerged as an important direction for diffusion efficiency by exploiting cross-timestep feature similarity in a lightweight, plug-and-play manner. Early U-Net-based\cite{unet} methods, such as DeepCache\cite{deepcache} and Faster Diffusion\cite{fasterdiffusion}, pioneered cross-timestep feature reuse, which was later extended to Diffusion Transformers (DiTs)\cite{dit} by FORA\cite{fora} and $\Delta$-DiT\cite{delta-dit}. More recent training-free methods, such as TeaCache\cite{teacache}, EasyCache\cite{easycache} and TaylorSeer\cite{taylorseers} have shown strong effectiveness in conventional multi-step diffusion, typically in the 30--50 step regime. A closely related work, FastCache\cite{fastcache}, uses a lightweight learnable linear layer to mitigate reuse error during multi-step inference. However, existing cache methods are largely tailored to standard multi-step sampling, where adjacent steps remain similar. This assumption breaks down in distilled few-step models, making naive feature reuse unreliable. How to effectively introduce caching into this regime without additional complex designs or training procedures remains largely unexplored.                                              
\section{Method}
\subsection{Preliminary: Distribution Matching Distillation}
\label{sec:prelim_dmd}
Our \emph{1.x-Distill} framework is built to overcome the limitations of distribution matching distillation. Therefore, we briefly introduce it as follows.

DMD\cite{dmd,dmd2} trains a few-step student generator $G_{\theta}$ to emulate the output distribution of a pre-trained diffusion model. 
This goal is formulated as minimizing the reverse Kullback--Leibler divergence between the student distribution $p_{\text{fake}}$ and the teacher-induced target distribution $p_{\text{real}}$:
\begin{equation}
    \label{eq:dmd_kl}
    \mathcal{L}_{\text{DMD}}(\theta)
    \;=\;
    \mathbb{E}_{x \sim p_\text{fake}}
    \Big[
    \mathrm{KL}\!\left(p_{\text{fake}}(x) \,\|\, p_{\text{real}}(x)\right)
    \Big].
\end{equation}
To train $G_{\theta}$ with this objective, the gradient of \cref{eq:dmd_kl} with respect to $\theta$ is calculated as:

\begin{equation}
\label{eq:dmd_grad}
\nabla_{\theta}\mathcal{L}_{\text{DMD}}
\;=\;
\mathbb{E}_{{t \sim \mathcal{U}},z}
\Big[-
\big(
s_{\text{real}}(x_t) -\; s_{\text{fake}}(x_t)
\big)
\frac{\partial {\hat x_0}} {\partial \theta}
\Big].
\end{equation}
where $\hat x_0$ is the denoising prediction of student generator $G_{\theta}$ and $x_t \sim q(x_t \mid \hat{x}_0,t)$ is the sample noised by perturbing the $\hat{x}_0$ according to the diffusion process at level $t \sim \mathcal{U}(0,1)$.
The score functions\cite{score-base-diff}
$s_{\text{real}}(x_t)\triangleq\nabla_{x_t}\log p_{\text{real}}(x_t)$ and
$s_{\text{fake}}(x_t)\triangleq\nabla_{x_t}\log p_{\text{fake}}(x_t)$
are vector fields that point toward higher-density regions of the corresponding distributions at noise level $t$. While the real score is estimated by the pretrained model itself, the fake score is estimated by a multi-step proxy that is dynamically updated to describe $p_{\text{fake}}$ with diffusion loss.
In \cref{eq:dmd_grad}, the difference $s_{\text{real}}(x_t)-s_{\text{fake}}(x_t)$ drives the student update by pushing its samples toward the teacher-induced target distribution.

\subsection{Controlling Guidance in Distribution Matching} 
\label{sec:cfg}

\begin{figure}[h]
  \centering
    \includegraphics[width=0.8\columnwidth]{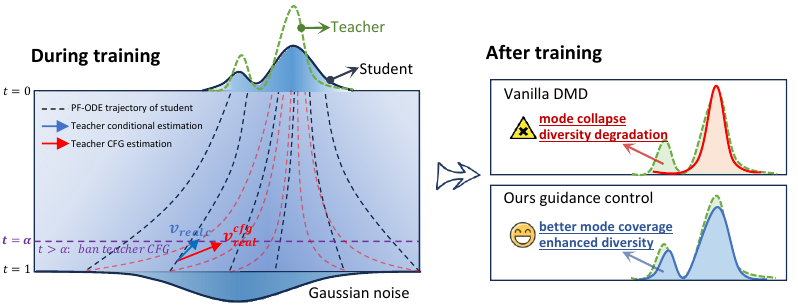} 
  \caption{\textbf{An illustration of the effect of the teacher’s CFG in distillation.} At the high-noise timestep $t$, teacher estimation with strong guidance \({\color{red}{v^\text{cfg}_\text{real}}}={\color{blue}{v_{\text{real},\text{c}}}}+(w-1)(v_{\text{real},\text{c}}-v_{\text{real},\emptyset})\) tends to drives the student to collapse prematurely toward dominant modes. We propose to disable teacher CFG at \(t\in(0,\alpha]\) during distribution matching, encouraging the student to cover more modes during early denoising trajectory.}
  \label{fig:cfg}
\end{figure}

Classifier-Free Guidance (CFG)\cite{cfg} is a pervasive component in diffusion inference, yet its role in distribution matching distillation has been largely under-discussed. We notice that in previous open-source DMD-like methods\cite{dmd2,tdm,senseflow}, the real score in \cref{eq:dmd_grad} is practically calculated with CFG under a strong guidance scale \(w\):
\begin{equation}
\label{eq:cfg}
\begin{aligned}
s_{\text{real}}(x_t)
&= s_{\text{real},\emptyset}(x_t)
   + w\bigl(s_{\text{real},c}(x_t)-s_{\text{real},\emptyset}(x_t)\bigr) \\
&= s_{\text{real},c}(x_t)
   + (w-1)\bigl(s_{\text{real},c}(x_t)-s_{\text{real},\emptyset}(x_t)\bigr).
\end{aligned}
\end{equation}
where $s_{\text{real},\emptyset}$ and $s_{\text{real},c}$ are the unconditional and conditional score estimation of the teacher model, respectively. This has also been noted in the a recent study\cite{decoupleddmd}, but we offer a different perspective in that overly strong guidance in the real score is an important driver of the \emph{mode collapse} commonly observed in DMD-like methods.

Along the denoising trajectory of a multi-step diffusion model, CFG critically affects the diversity and fidelity trade-off. A higher guidance scale \(w\) improves prompt adherence and fine details, while weaker guidance increases sample diversity. This mechanism also appears in DMD training. 
As shown in \cref{fig:cfg}, matching a strongly guided real score yields overly biased supervision. 
In high-noise regimes, the biased target forces the student to match a mode-seeking score field rather than the full data distribution.
As a result, the student is encouraged to collapse toward a few dominant modes early in the denoising trajectory, leading to severe diversity degradation.

A naïve remedy is to globally reduce the teacher guidance scale during distillation, but this substantially degrades quality by weakening the visual constraints for detail synthesis.
We find that applying CFG at early timesteps more directly harms diversity, a phenomenon that has also been observed in multi-step diffusion sampling~\cite{applycfg}. Therefore, we control the teacher guidance in a timestep-aware manner when constructing the real-score target:

\begin{equation}
\label{eq:cfg_ccd}
s_{\text{real}}(x_t)=
\begin{cases}
s_{\text{real},\emptyset}(x_t)\;+\;w\!\left(s_{\text{real},c}(x_t)-s_{\text{real},\emptyset}(x_t)\right), & t\in(0,\alpha]\\[4pt]
s_{\text{real},c}(x_t), & t\in(\alpha,1]
\end{cases}
\end{equation}

Following Eq.~\eqref{eq:cfg_ccd}, we disable CFG in real score estimation for early timesteps \(t\in(\alpha,1]\) and use the fully conditional score \(s_{\text{real},c}(x_t)\) instead, encouraging the student to learn richer coarse structures and cover more modes during early denoising trajectory. For mid-to-low noise level at \(t\in(0,\alpha]\), it is necessary to maintain strong guidance to preserve prompt alignment and fine details. This simple modification retains the DMD framework, yet significantly improves diversity without sacrificing fidelity.

\begin{figure}[t]
  \centering
     \includegraphics[width=\columnwidth]{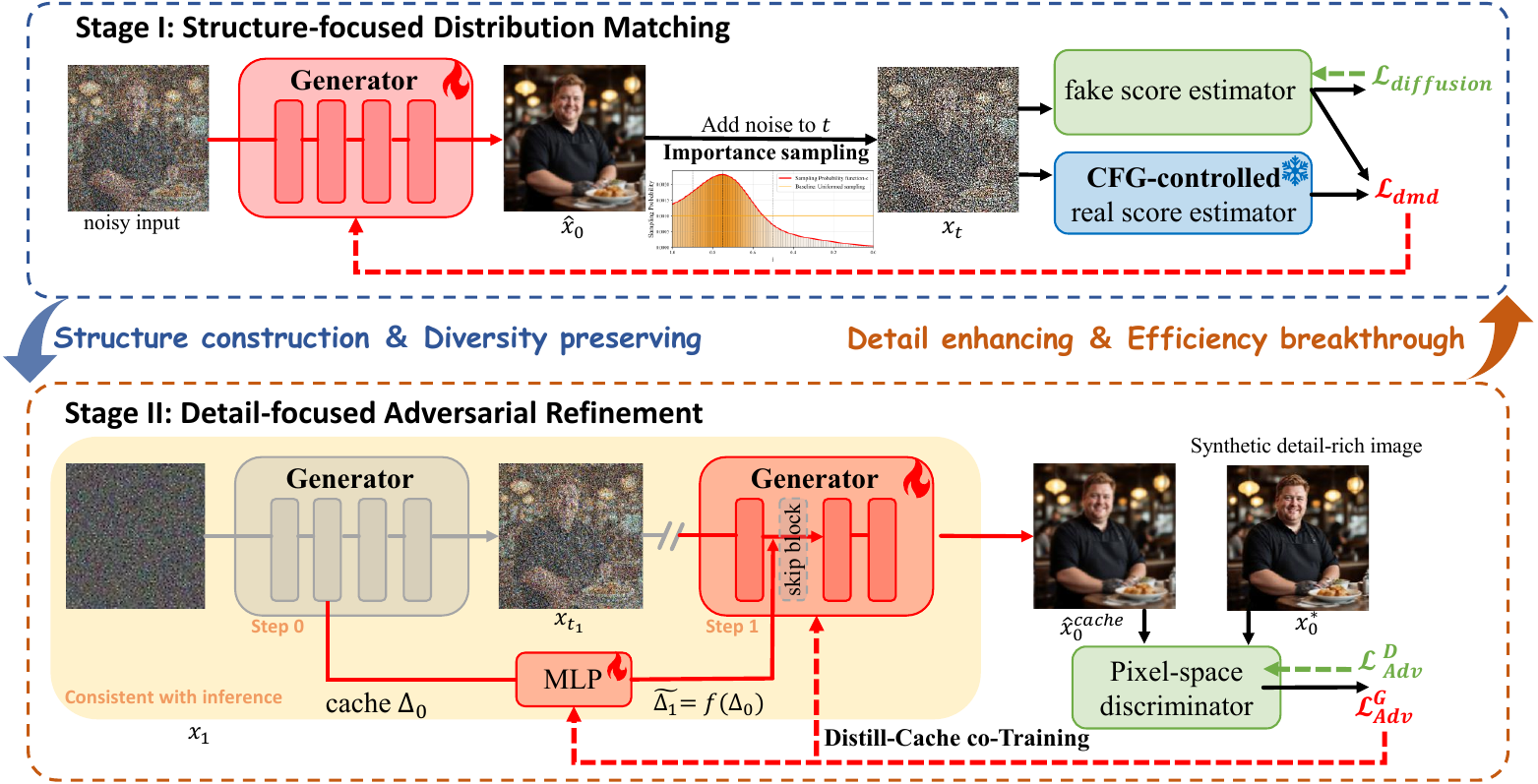} 
     
     \caption{\textbf{Overview of \emph{1.x-Distill}}. Our guidance control (\cref{sec:cfg}) and cache design (\cref{sec:cache}) are both constructed in the two-stage framework (\cref{sec:SFD}). \textbf{Stage~I}: Train the generator with DMD loss. Within DMD framework, we apply \emph{importance sampling} on diffusion timestep \(t\), and \emph{control the guidance} according to sampled \(t\) when compute the real score. \textbf{Stage~II}: Train the generator with pixel-space adversarial loss. Our GAN framework produces \(\hat x_0\) along generator inference path, which naturally incorporates block-cache design. The generator and MLP module are jointly optimized. 
     }

  \label{fig:overview}
\end{figure}
\subsection{Stagewise Focused Distillation}
\label{sec:SFD}

Extreme 2-step distillation forces each step to handle both global structure and fine details, making a single uniform objective misaligned with learning dynamics. We propose \textbf{Stagewise Focused Distillation}, a two-stage framework with \emph{Structure-focused Distribution Matching} for robust structure and \emph{Detail-focused Adversarial Refinement} for fine details.

\begin{figure}[h]
  \centering
    \includegraphics[width=\columnwidth]{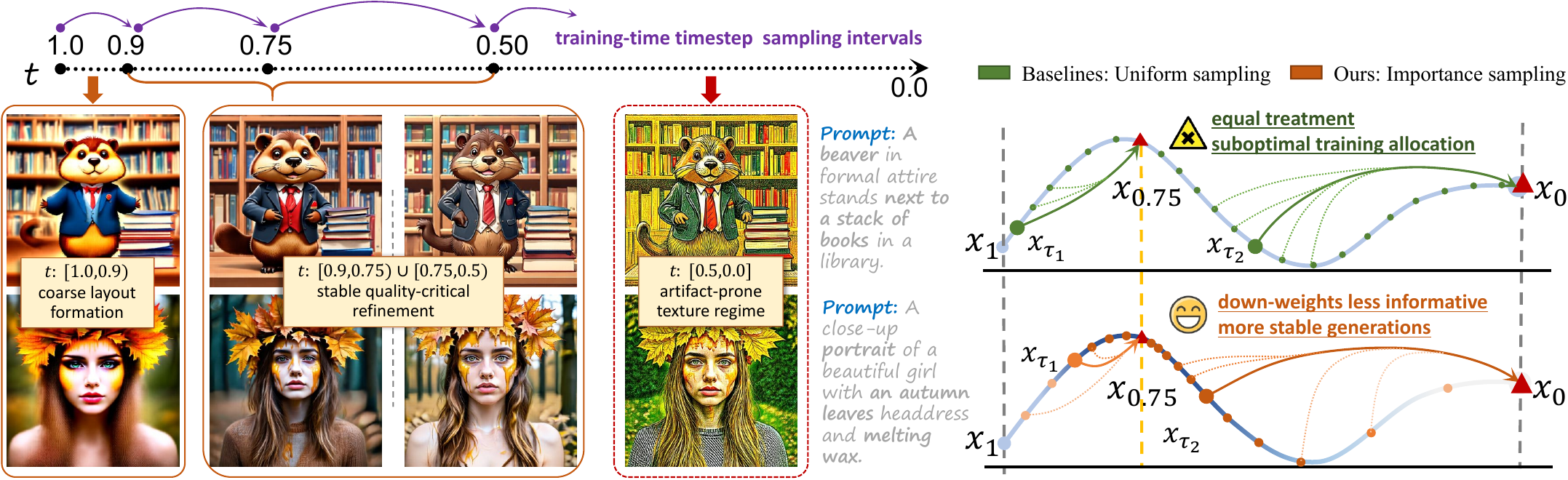} 
  \caption{\textbf{Importance sampling in Stage~I}. \emph{Left}: Under teacher scheduler (shift=3.0), we split timesteps from 1.0 to 0.0 into four windows to probe their effects.  
  \emph{Right}: Uniform sampling treats all timesteps equally, while our importance sampling down-weights less informative ones and concentrates training on the more reliable region.}
  \label{fig:method_win}
\end{figure}
\subsubsection{Stage~I: Structure-focused Distribution Matching}
\label{sec:stage1}

Conventional distribution matching is suboptimal in the extreme few-step regime, where stable optimization becomes much more difficult. As shown in \cref{fig:method_win}, excessive updates from low-noise timesteps \((t\in(0,0.5))\) are dominated by local texture perturbations, leading to over-sharpened images and abnormal color artifacts. This indicates that uniform timestep sampling misallocates training effort in Stage~I. To address this, we design a \emph{importance timestep sampling} strategy for the structure-focused stage. Under teacher scheduler setting (shift=3.0), the sampling probability peaks around \(t= 0.75\) and decays rapidly when \(t<0.5\), shifting optimization away from low-noise texture corrections and toward structurally informative timesteps, following the probability curve in \cref{fig:overview}.

\subsubsection{Stage~II: Detail-focused Adversarial Refinement}
After Stage~I, the student already produces structurally plausible two-step samples with stable semantics. We therefore introduce a pixel-space GAN~\cite{gan} objective in Stage~II to further refine fine-grained details:

\begin{equation}
\label{eq:gan}
\begin{aligned}
\mathcal{L}_{\mathrm{Adv}}^{G}
&=
\mathbb{E}_{\hat{x}_0}
\left[
-\log D\left(V(\hat{x}_0)\right)
\right],\\
\mathcal{L}_{\mathrm{Adv}}^{D}
&=
\mathbb{E}_{x^{\star}_0}
\left[
-\log D\left(V(x^{\star}_0)\right)
\right]
+
\mathbb{E}_{\hat{x}_0}
\left[
\log D\left(V(\hat{x}_0)\right)
\right],
\end{aligned}
\end{equation}
where $V$ denotes the VAE decoder and $D$ is the pixel-space discriminator. Prior methods\cite{dmd2,senseflow} jointly optimize the generator with the DMD loss~\cref{eq:dmd_grad} and the GAN loss, generating samples \(\hat{x}_0\) via single-step prediction from randomly sampled noise levels \((t\in(0,1])\).
Such training introduces large variation in generator outputs, making discriminator optimization unstable.
In contrast, our GAN framework maintains training-inference consistency. We generate $\hat x_0$ along the few-step inference path and forward propagate the generator in the last step to focus on refinement without disrupting the structure learned in Stage I. 
We further simplify the construction of real samples. The ``real'' images \(x^{\star}_0\) in our adversarial training are generated from the same noise by a multi-step model. Multi-step synthetic \(x^{\star}_0\) typically exhibit richer details while remaining more structurally consistent with the distilled distribution. Consequently, our formulation removes the reliance on high-quality image datasets and reducing the domain gap between real and generated images that can otherwise bias detail refinement.
For the discriminator, we follow the architecture in~\cite{hypir}. A frozen ConvNeXt\cite{convnext} backbone is used to extract fine-grained features, followed by a trainable classification head, which empirically performs well for detail-oriented refinement tasks.

\subsection{Caching for Distilled Model}
\begin{figure}[h]
  \centering
    \includegraphics[width=0.97\columnwidth]{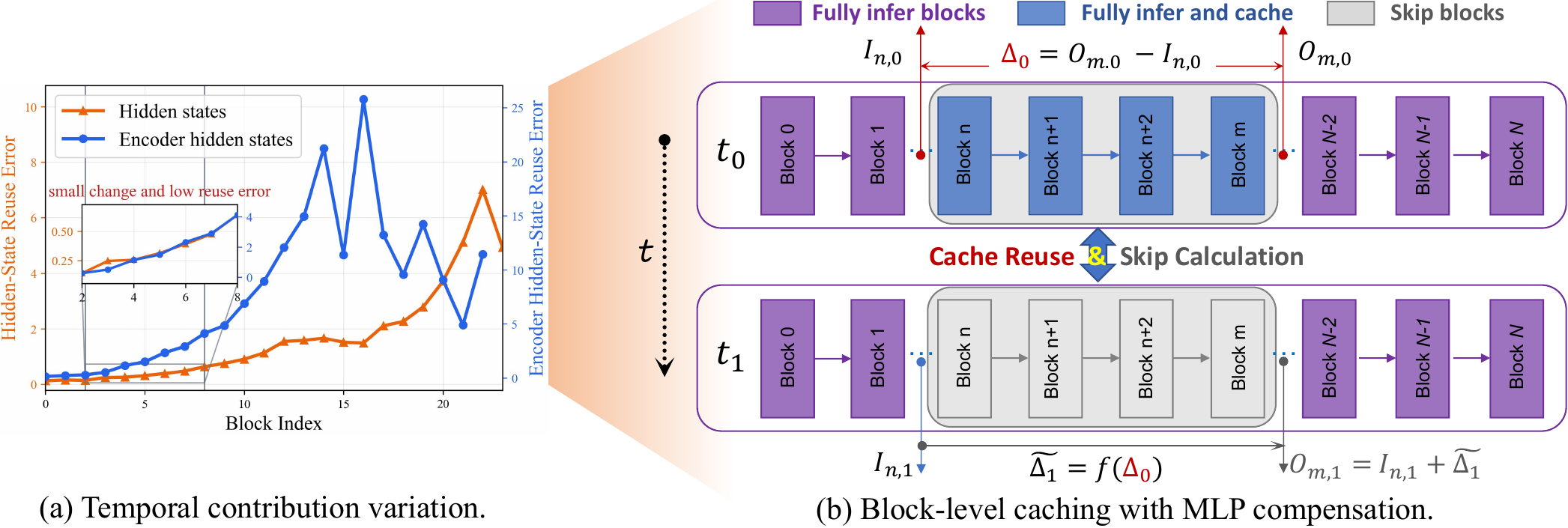} 
    \caption{\textbf{Caching for a distilled 2-step student.} 
    \label{fig:method_cache}
    \textbf{(a)} We measure block-wise reuse error as the contribution change across adjacent steps on \emph{SD3-M}, {\color{red}$e_n=\lVert \Delta_{n,t+1}-\Delta_{n,t}\rVert_1$}, where $\Delta_{n,t}=O_{n,t}-I_{n,t}$. Early blocks exhibit consistently small $e_n$, indicating strong temporal redundancy and low reuse error. \textbf{(b)} Leveraging this property, we cache the contribution of a block segment $[n,m]$ at step $t_0$, {\color{red}$\Delta_0=O_{m,0}-I_{n,0}$}, skip the segment at $t_1$, and recover the output via $\hat{O}_{m,1}=I_{n,1}+f({\color{red}\Delta_0})$.}
    
  \label{fig:cache}
\end{figure}
Our SFD achieves high-quality 2-step sampling, while direct 1-step distillation still degrades quality. To eliminate redundant computation in full per-iteration computation, we introduce \emph{block-level caching} into the 2-step DiT-based student, pushing efficiency further and achieving 1.x-NFE inference.

\subsubsection{Cache Design for 2-step Student}
\label{sec:cache}
We implement cache-accelerated inference through block-level feature reuse across consecutive denoising steps. Suppose the model is fully evaluated at step \(t\), and a block segment \([n,m]\) is skipped at step \(t+1\). Let \(I_{n,t}\) and \(O_{m,t}\) denote the input of block \(n\) and the output of block \(m\), respectively. We cache the block contribution
\[
\Delta_t = O_{m,t} - I_{n,t},
\]
and directly reuse it to bypass the skipped computation at the next step. To reduce the resulting reuse error, we introduce a \textbf{learnable error-compensation module} \(f(\cdot)\), implemented as a lightweight residual MLP, and predict the reused contribution as
\[
\tilde{\Delta}_{t+1}=f(\Delta_t),
\qquad
\hat{O}_{m,t+1}=I_{n,t+1}+\tilde{\Delta}_{t+1}.
\]
Since \(f(\cdot)\) is negligible compared with the skipped DiT blocks, this design adds little overhead while substantially reducing reuse error. In our 2-step setting, the first step is fully computed and the second step reuses the predicted block contribution.

\subsubsection{Distill--Cache co-Training}
\label{sec:co_training}
Under our SFD framework, Stage II naturally supports cache recovery training, as the adversarial refinement strictly aligns with the inference pipeline. With caching enabled and the correction module $f$ inserted before Stage II, the adversarial objective directly supervises cache-accelerated inference and helps recover from cache-induced distortions.
Denote by $\hat{x}_0^{\mathrm{cache}}$ the image decoded from the student output produced by the cache-accelerated second step. We optimize the adversarial objective: 
\begin{equation}
\label{eq:cache_gan_obj}
\min_{\theta,\,\phi}\;\mathcal{L}_{\mathrm{Adv}}^{G}
=
\min_{\theta,\,\phi}\;\mathbb{E}_{\hat{x}_0^{\mathrm{cache}}}
\left[
-\log D\left(V(\hat{x}_0^{\mathrm{cache}})\right)
\right],
\end{equation}
where $\theta$ denotes the parameters of the student backbone $G_{\theta}$ and $\phi$ denotes the parameters of the correction module $f$. In practice, we first freeze $\theta$ and warm up $\phi$ for a few iterations. We then jointly optimize detail enhancement and cache recovery under the same adversarial supervision. Notably, we require no feature-level alignment loss, as pixel-level adversarial supervision alone compensates reuse error, restores image quality, and enables stable fractional-step inference.
\section{Experiment}

\subsection{Experimental Setup}
\subsubsection{Settings}

We apply \textbf{\emph{1.x-Distill}} to two representative DiT-based text-to-image models, \emph{SD3-Medium} (2B) and \emph{SD3.5-Large} (8B)\cite{sd3}. To provide a clear comparison of acceleration, we define the \emph{effective NFE} (number of function evaluations) as the ratio of fully computed DiT blocks during student sampling to the total number of blocks in the original model. For \emph{SD3-Medium} with 24 DiT blocks, skipping layers 3--8 in the second denoising step yields an effective NFE of 1.75, while skipping layers 3--10 further reduces it to 1.67. For \emph{SD3.5-Large} with 38 DiT blocks, skipping layers 3--12 in the second step yields an effective NFE of 1.74. We also report 4-step results of SFD without caching for direct comparison with prior 4-step methods. Since our adversarial training does not rely on external image datasets, we use only prompt data from JourneyDB\cite{journeydb} throughout training. More implementation and training details of our method are provided in the supplementary material.

\subsubsection{Baselines}

We compare our method against all publicly available few-step checkpoints of \emph{SD3-Medium} and \emph{SD3.5-Large}, including trajectory- and distribution-based methods like Hyper-SD\cite{hypersd}, PCM\cite{pcm}, Flash\cite{flashdiff}, LADD(Turbo)\cite{ladd} and TDM\cite{tdm}. 
Since most open-source models do not directly support 2-step inference, for fair comparison we also try our best to implement the 2-step results of representative distribution matching methods, including DMD2\cite{dmd2} and TDM. 


\definecolor{headviolet}{RGB}{252,228,236}   
\definecolor{headpink}{RGB}{222,233,249}     
\definecolor{oursblue}{RGB}{156,182,229}     
\definecolor{oursbluefill}{RGB}{222,233,249} 
\definecolor{incblue}{RGB}{45,92,195}        
\definecolor{redhi}{RGB}{200,0,0}

\begin{table*}[h]
\centering
\caption{\textbf{Quantitative comparison on COCO-10K.} $^{*}$ indicates results reproduced by us due to missing official checkpoints. FID is computed between teacher samples and student samples. Img-free indicates the method does not use external real-image datasets during training.}
\label{tab:main_results}
\small
\setlength{\tabcolsep}{4.5pt}
\renewcommand{\arraystretch}{1.2}
\resizebox{0.97\linewidth}{!}{
\begin{tabular}{l|cc|cccccc|c}
\toprule
\multicolumn{1}{c}{\textbf{Method}} &
\multicolumn{1}{c}{\cellcolor{headviolet}\textbf{Step}} &
\multicolumn{1}{c}{\cellcolor{headviolet}\textbf{\#NFE}} &
\multicolumn{1}{c}{\cellcolor{headpink}FID\cite{fid}$\downarrow$} &
\multicolumn{1}{c}{\cellcolor{headpink}CLIP\cite{clipscore}$\uparrow$} &
\multicolumn{1}{c}{\cellcolor{headpink}AS\cite{laion5b}$\uparrow$} &
\multicolumn{1}{c}{\cellcolor{headpink}PS\cite{pickscore}$\uparrow$} &
\multicolumn{1}{c}{\cellcolor{headpink}IR\cite{ir}$\uparrow$} &
\multicolumn{1}{c}{\cellcolor{headpink}HPSv2\cite{hpsv2}$\uparrow$} &
\multicolumn{1}{c}{\textbf{Img-free}} \\
\midrule

\multicolumn{10}{c}{\textbf{Stable Diffusion 3 Medium} $1024{\times}1024$} \\
\midrule
Base Model          & 28 & $28{\times}2$ & -- & 0.3176 & 5.6348 & 22.5554 & 1.0429 & 30.7197 & \xmark \\
Hyper-SD\cite{hypersd}        & 4  & $4{\times}2$  & 15.5475 & 0.3127 & 4.9582 & 21.6407 & 0.7543 & 28.7578 & \xmark \\
PCM\cite{pcm}                 & 4  & 4             & 17.5605 & 0.3102 & \underline{5.7743} & 22.0690 & 0.6715 & 29.0864 & \xmark \\
Flash\cite{flashdiff}         & 4  & 4             & 15.6443 & \textbf{0.3166} & 5.5485 & 22.3879 & 0.8938 & 29.4835 & \xmark \\
DMD2$^{*}$\cite{dmd2}         & 4  & 4             & 14.7125 & 0.3122 & 5.4632 & 22.4120 & 0.9981 & 31.0152 & \xmark \\
TDM\cite{tdm}                 & 4  & 4             & \underline{14.6424} & 0.3128 & 5.5494 & \underline{22.4681} & \underline{1.0021} & \underline{31.7512} & \cmark \\
\rowcolor{gray!15}
Ours-SFD                      & 4  & 4             & \textbf{14.1349} & \underline{0.3149} & \textbf{5.9197} & \textbf{22.8155} & \textbf{1.1196} & \textbf{32.5337} & \cmark \\
{\color{incblue}\scriptsize $\Delta$ (vs best baseline)} &
{\color{incblue}\scriptsize --} & {\color{incblue}\scriptsize --} &
{\color{incblue}\scriptsize -0.5075} &
{\color{incblue}\scriptsize --} &
{\color{incblue}\scriptsize +0.1454} &
{\color{incblue}\scriptsize +0.2601} &
{\color{incblue}\scriptsize +0.0767} &
{\color{incblue}\scriptsize +0.7825} &
{\color{incblue}\scriptsize --} \\
\arrayrulecolor{gray!60}\midrule\arrayrulecolor{black}

PCM                           & 2  & 2             & 41.6561 & 0.3077 & 5.1325 & 20.9493 & 0.2011 & 24.8431 & \xmark \\
TDM$^{*}$                     & 2  & 2             & 19.3005 & 0.3186 & \underline{5.1441} & \underline{22.4756} & \underline{1.1101} & 31.4725 & \cmark \\
\rowcolor{gray!15}
\emph{1.x-Distill}-slow        & 2  & {\color{redhi}\textbf{1.75}} & \textbf{15.7863} &
\textbf{0.3208} & \textbf{5.1844} & \textbf{22.5161} & \textbf{1.1312} & \textbf{32.2550} & \cmark \\
{\color{incblue}\scriptsize $\Delta$} &
{\color{incblue}\scriptsize --} & {\color{incblue}\scriptsize --} &
{\color{incblue}\scriptsize -3.5142} &
{\color{incblue}\scriptsize +0.0022} &
{\color{incblue}\scriptsize +0.0403} &
{\color{incblue}\scriptsize +0.0405} &
{\color{incblue}\scriptsize +0.0211} &
{\color{incblue}\scriptsize +0.7825} &
{\color{incblue}\scriptsize --} \\
\rowcolor{gray!15}
\emph{1.x-Distill}-fast        & 2  & {\color{redhi}\textbf{1.67}} & \underline{16.7179} &
\underline{0.3204} & 5.1206 & 22.3342 & 1.0673 & \underline{31.6850} & \cmark \\

\midrule
\multicolumn{10}{c}{\textbf{Stable Diffusion 3.5 Large} $1024{\times}1024$} \\
\midrule
Base Model                    & 28 & $28{\times}2$ & - & 0.3196 & 5.9178 & 22.5994 & 1.0641 & 31.1081 & \xmark \\
Turbo\cite{ladd}        & 4  & 4             & \textbf{15.3123} & \underline{0.3161} & \textbf{6.1308} & \underline{22.7418} & \underline{0.9288} & \underline{30.4127} & \xmark \\
\rowcolor{gray!15}
Ours-SFD                      & 4  & 4             & \underline{17.3588} & \textbf{0.3187} & \underline{5.9939} & \textbf{22.9046} & \textbf{1.2011} & \textbf{32.9020} & \cmark \\
{\color{incblue}\scriptsize $\Delta$} &
{\color{incblue}\scriptsize --} & {\color{incblue}\scriptsize --} &
{\color{incblue}\scriptsize --} &
{\color{incblue}\scriptsize +0.0026} &
{\color{incblue}\scriptsize --} &
{\color{incblue}\scriptsize +0.1628} &
{\color{incblue}\scriptsize +0.1370} &
{\color{incblue}\scriptsize +1.7939} &
{\color{incblue}\scriptsize --} \\
\arrayrulecolor{gray!60}\midrule\arrayrulecolor{black}

TDM$^{*}$                     & 2  & 2             & \underline{26.8084} & \textbf{0.3224} & \underline{5.3110} & \underline{22.1424} & \underline{0.9307} & \underline{28.4919} & \cmark \\
\rowcolor{gray!15}
\emph{1.x-Distill}             & 2  & {\color{redhi}\textbf{1.74}} & \textbf{22.0545} &
\underline{0.3191} & \textbf{5.7976} & \textbf{22.7963} & \textbf{1.1463} & \textbf{32.0059} & \cmark \\
{\color{incblue}\scriptsize $\Delta$} &
{\color{incblue}\scriptsize --} & {\color{incblue}\scriptsize --} &
{\color{incblue}\scriptsize -4.7539} &
{\color{incblue}\scriptsize --} &
{\color{incblue}\scriptsize +0.4866} &
{\color{incblue}\scriptsize +0.6539} &
{\color{incblue}\scriptsize +0.2156} &
{\color{incblue}\scriptsize +3.5140} &
{\color{incblue}\scriptsize --} \\

\bottomrule
\end{tabular}
}
\vspace{-0.4cm}
\end{table*}

\subsection{Main Results}
\subsubsection{Quantitative Comparison}
\begin{figure}[t]
  \centering
    \includegraphics[width=0.98\columnwidth]{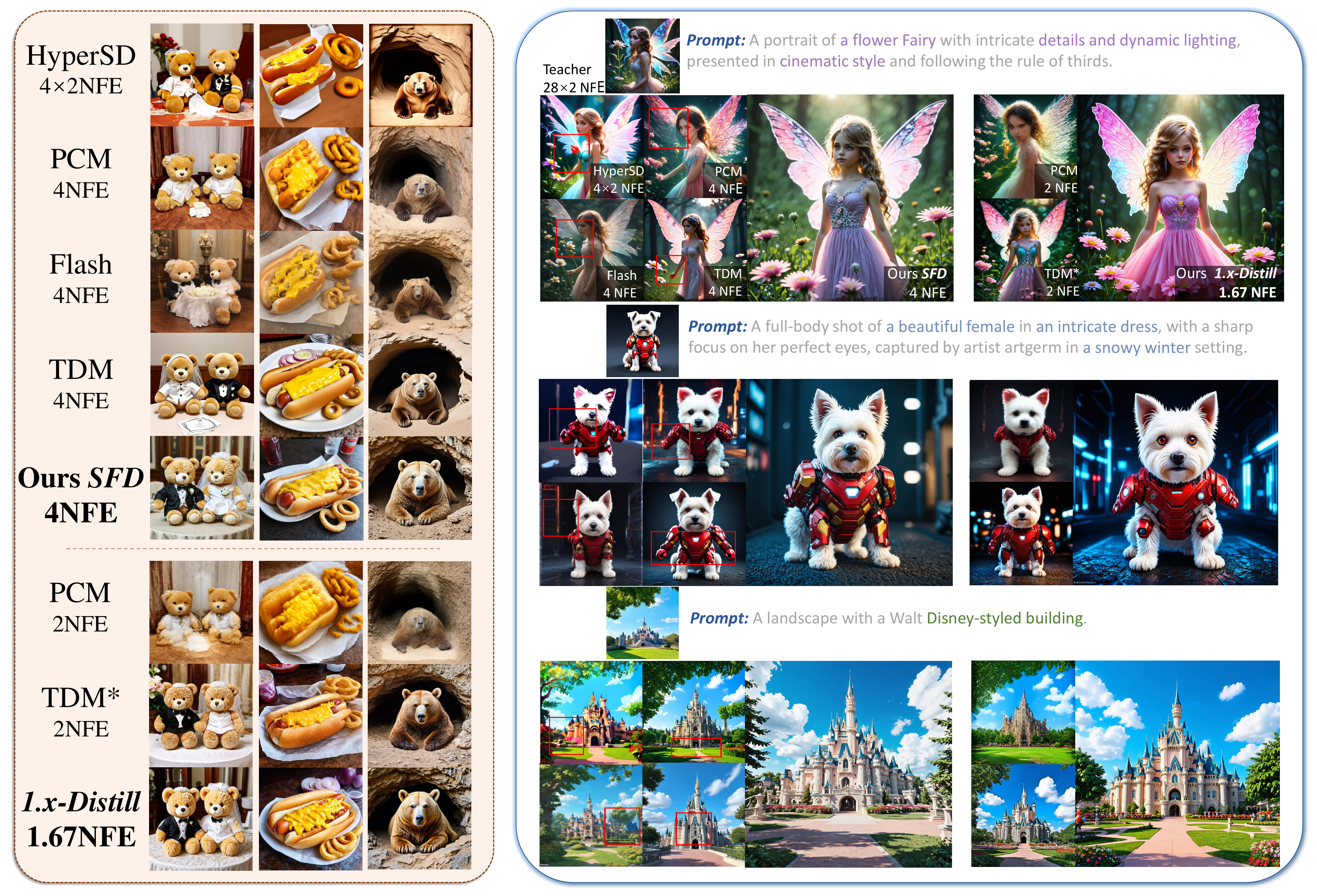} 
  \caption{\textbf{Qualitative comparison on \emph{SD3-Medium}}. Since most open-source baselines only provide 4-step checkpoints, we first compare all methods at 4 steps for fairness. Our \textbf{SFD} already produces clearer and more appealing images, while other methods often show generation failures, color shifts, blur, or degraded aesthetics (\textcolor{red}{red boxes}). When pushed to 2 steps, \textbf{\emph{1.x-Distill}} still clearly surpasses all baselines.}
  \label{fig:comparison}
\end{figure}

\definecolor{headviolet}{RGB}{252,228,236}  
\definecolor{headpink}{RGB}{222,233,249}    
\definecolor{redhi}{RGB}{200,0,0}

\begin{table*}[h]
\centering
\small

\begin{minipage}[t]{0.58\textwidth}
\centering
\caption{Quantitative evaluation on DPG-Bench of our \emph{1.x-Distill} model against its multi-step teacher.}
\label{tab:bpgbench}
\setlength{\tabcolsep}{3pt}
\renewcommand{\arraystretch}{1.2}
\resizebox{\linewidth}{!}{
\begin{tabular}{l|c|c|ccccc}
\toprule
\multicolumn{1}{c}{\textbf{Model}} &
\multicolumn{1}{c}{\cellcolor{headviolet}\textbf{\#NFE}} &
\multicolumn{1}{c}{\cellcolor{headpink}\textbf{Overall$\uparrow$}} &
\multicolumn{1}{c}{\cellcolor{headpink}Global} &
\multicolumn{1}{c}{\cellcolor{headpink}Entity} &
\multicolumn{1}{c}{\cellcolor{headpink}Attribute} &
\multicolumn{1}{c}{\cellcolor{headpink}Relation} &
\multicolumn{1}{c}{\cellcolor{headpink}Other} \\
\midrule
SD3-M  & $28{\times}2$ & 85.46 & 92.01 & 89.07 & 89.88 & 90.46 & 91.67 \\
Ours   & {\color{redhi}\textbf{1.75}} & \textbf{86.13} & 89.97 & \textbf{92.26} & 89.41 & \textbf{90.77} & \textbf{92.01} \\
\arrayrulecolor{gray!60}\midrule\arrayrulecolor{black}
SD3.5-L & $28{\times}2$ & 84.74 & 90.02 & 89.67 & 90.97 & 89.70 & 89.53 \\
Ours    & {\color{redhi}\textbf{1.74}} & \textbf{85.11} & 89.78 & \textbf{90.79} & 89.49 & \textbf{92.21} & \textbf{89.66} \\
\bottomrule
\end{tabular}
}
\end{minipage}
\hfill
\begin{minipage}[t]{0.38\textwidth}
\centering
\small
\caption{Quantitative evaluation of diversity on COCO-1K using LPIPS.}
\label{tab:diversity}
\setlength{\tabcolsep}{6pt}
\renewcommand{\arraystretch}{1.15}
\resizebox{0.95\linewidth}{!}{
{\scriptsize
\begin{tabular}{l|c|c}
\toprule
\multicolumn{1}{c}{\textbf{Method}} &
\multicolumn{1}{c}{\cellcolor{headviolet}\textbf{\#NFE}} &
\multicolumn{1}{c}{\cellcolor{headpink}\textbf{LPIPS$\uparrow$}} \\
\midrule
SD3-M  & $28{\times}2$ & 0.6594 \\
Flash  & 4             & 0.6161 \\
TDM    & 4             & 0.6297 \\
\rowcolor{gray!15}
Ours   & {\color{redhi}\textbf{1.75}} & \textbf{0.6432} \\
\bottomrule
\end{tabular}
}
}
\end{minipage}
\end{table*}
Following prior work\cite{dmd,dmdx}, we conduct our evaluation on 10K prompts from COCO-2014\cite{coco2014}. We report FID\cite{fid} for distribution fidelity, CLIP Score\cite{clipscore} for prompt alignment and commonly used human-preference metrics including Pick Score\cite{pickscore}, Aesthetic Score\cite{laion5b}, HPSv2\cite{hpsv2}, and ImageReward\cite{ir}.
As shown in \cref{tab:main_results}, \emph{1.x-Distill} achieves a strong quality--efficiency trade-off on both \emph{SD3-M} and \emph{SD3.5-L}. Before caching, our \textbf{SFD} already performs strongly at 4 step, achieving the best preference scores on \emph{SD3-M}. After enabling block caching, the advantage becomes clearer in the extreme few-step regime. On \emph{SD3-M}, \textbf{\emph{1.x-Distill}}--slow surpasses the strongest reproduced baseline TDM at a lower effective NFE (1.75 vs.\ 2), and even outperforms all existing 4-NFE methods on most quality metrics. Increasing the cache ratio, \textbf{\emph{1.x-Distill}}--fast further reduces the effective NFE to 1.67 with only a modest performance drop.
We further evaluate on DPG-Bench\cite{dpgbench}, a popular text-to-image benchmark, to comprehensively assess our models under complex prompts. As shown in \cref{tab:bpgbench}, our distilled models outperform the original multi-step teachers in overall score under aggressive step compression.

In addition, we evaluate sampling diversity using LPIPS\cite{lpips}. For each prompt, we generate four samples with different seeds and compute pairwise LPIPS distances, averaged over 1K COCO-2014 prompts. Results in \cref{tab:diversity} show that our method achieves substantially higher diversity than prior distribution-matching baselines (Flash and TDM).

\subsubsection{Qualitative Comparison}
In addition to quantitative comparisons, we present qualitative results in \cref{fig:comparison}. 
Across a wide range of prompts, \emph{1.x-Distill} consistently produces visually superior images compared to prior methods. 
Remarkably, even under the extremely low compute budget (1.67 NFE), our distilled model preserves coherent global structure while generating rich and realistic fine details, even surpassing the teacher model. 
Besides, see the visual comparison of the diversity in the supplementary material.

\subsubsection{User Study}
\begin{figure}[h]
  \centering
    \includegraphics[width=\columnwidth]{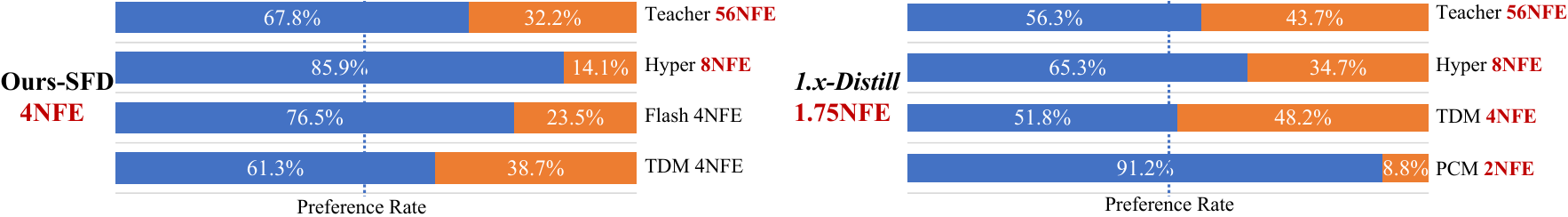} 
  \caption{User study: Comparing images generated by \emph{1.x-Distill} with other models.}
  \label{fig:user_study}
\end{figure}

We conduct a user study to assess perceptual quality and prompt alignment. 
20 human raters compare images generated by our method against strong few-step baselines on 3,200 prompts of 4 styles from HPSv2. 
The results in \cref{fig:user_study} show a clear human preference for \emph{1.x-Distill}.

\subsection{Ablation Studies}

We perform comprehensive ablation studies to validate the effectiveness of each component and identify optimal design choices. Unless otherwise noted, all studies are performed in \emph{SD3-Medium} at $1024\times1024$.

\begin{center}
\includegraphics[width=0.9\columnwidth]{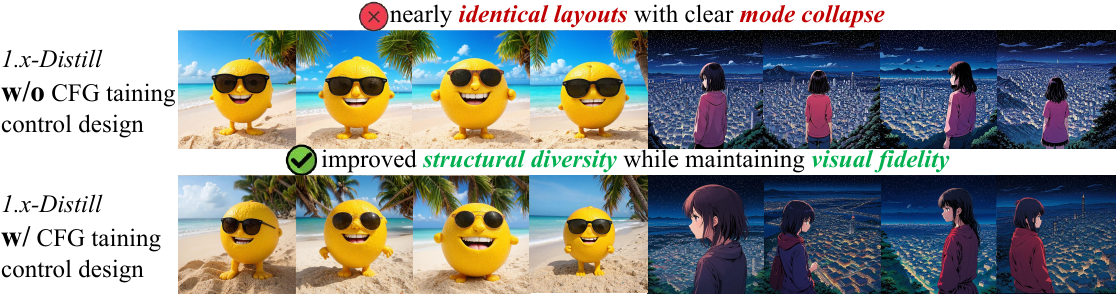}
\captionof{figure}{Effect of guidance control strategy on sample diversity.}
\label{fig:ablation_cfg}
\end{center}

\subsubsection{Effect of Guidance Control}

To validate our guidance control strategy, we compare the sampling results of \emph{1.x-Distill} with and without it under a unified guidance scale $w=7.0$. 
As shown in \cref{fig:ablation_cfg}, enabling guidance control produces more diverse structural layouts while preserving comparable image quality.
As discussed in \cref{sec:cfg}, completely disabling the teacher CFG in the mid-to-low noise regime may harm distillation results. 
We further vary the threshold $\alpha$ to identify the optimal control boundary. 
As shown in \cref{fig:ablation_cfg}, when $\alpha < 0.92$, the distilled model exhibits clear degradation in quality metrics. 
This is because the student increasingly relies on mid-to-low noise timesteps to learn perceptual quality rather than structural diversity.
We therefore set $a = 0.94$ in our \emph{1.x-Distill}. 

\noindent
\begin{minipage}[t]{0.48\columnwidth}
    \centering
    \includegraphics[width=\linewidth]{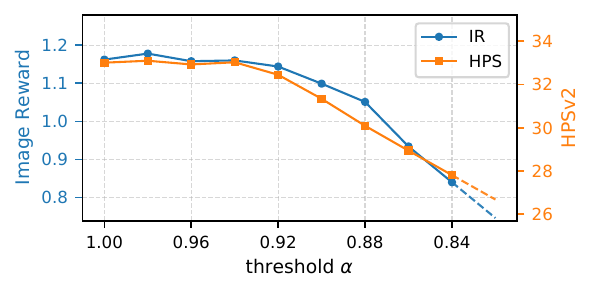}
    \captionof{figure}{Effect of control threshold \(\alpha\).}
    \label{fig:ir_hps_vs_a}
\end{minipage}
\hfill
\begin{minipage}[t]{0.48\columnwidth}
    \centering
    \includegraphics[width=\linewidth]{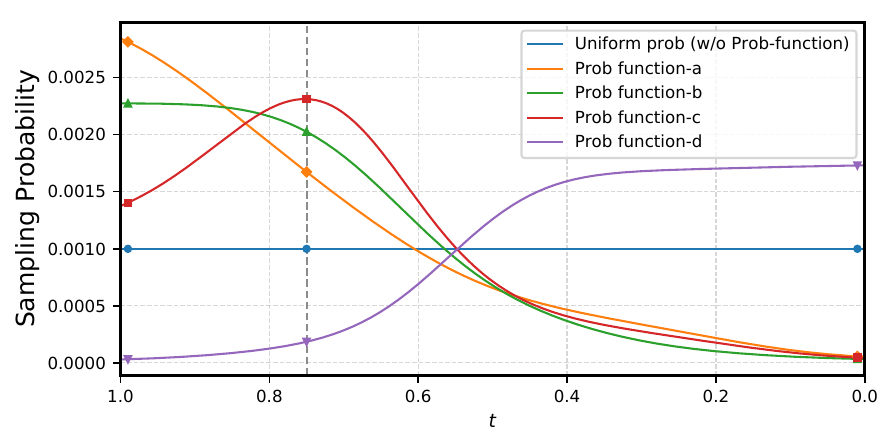}
    \captionof{figure}{Ablation of sampling probability.}
    \label{fig:ablation_prob_curves}
\end{minipage}

\definecolor{headblue}{RGB}{222,233,249}
\definecolor{redhi}{RGB}{200,0,0}
\definecolor{headpink}{RGB}{222,233,249}
\begin{table}[h]
\centering
\caption{\textbf{Ablations of Stagewise Focused Distillation (SFD).} \textit{Top}: Stage~I timestep sampling ablations where our strategy (c in \cref{fig:ablation_prob_curves}) outperforms uniform sampling. \textit{Bottom}: Stage~II further improves detail fidelity over Stage~I. All settings use the same training iterations with controlled guidance.}
\label{tab:ablation_sfd}
\scriptsize
\setlength{\tabcolsep}{3.5pt}
\renewcommand{\arraystretch}{1.2}
\resizebox{0.78\columnwidth}{!}{
\begin{tabular}{cc|ccccc}
\toprule
\multicolumn{1}{c}{\textbf{Stage I}} &
\multicolumn{1}{c}{\textbf{Stage II}} &
\multicolumn{1}{c}{\cellcolor{headpink}CLIP$\uparrow$} &
\multicolumn{1}{c}{\cellcolor{headpink}AS$\uparrow$} &
\multicolumn{1}{c}{\cellcolor{headpink}PS$\uparrow$} &
\multicolumn{1}{c}{\cellcolor{headpink}IR$\uparrow$} &
\multicolumn{1}{c}{\cellcolor{headpink}HPSv2$\uparrow$} \\
\midrule
\checkmark: Uniform sampling & \xmark & 0.3154 & 5.1432 & 22.4546 & 1.1081 & 31.4725 \\
\checkmark: a & \xmark & \underline{0.3161} & 5.6044 & 22.4337 & 1.1170 & 31.8419 \\
\checkmark: b & \xmark & 0.3155 & 5.5442 & 22.3769 & 1.1020 & 31.8944 \\
\rowcolor{gray!15}
{\color{redhi}\textbf{\checkmark: c}} & \xmark & 0.3159 & \underline{5.6210} & \underline{22.4694} & 1.1226 & \underline{32.0208} \\
\checkmark: d & \xmark & 0.3127 & 4.6196 & 22.4185 & \underline{1.1419} & 30.2768 \\
\arrayrulecolor{gray!60}\midrule\arrayrulecolor{black}
\xmark & \checkmark & 0.2144 & 3.3287 & 13.1621 & 0.1352 & 19.8334 \\
\rowcolor{gray!15}
{\color{redhi}\textbf{\checkmark: c}} & \color{redhi}\checkmark & \textbf{0.3184} & \textbf{5.7485} & \textbf{22.6995} & \textbf{1.1601} & \textbf{33.0293} \\
\bottomrule
\end{tabular}
}
\vspace{-0.5cm}
\end{table}

\subsubsection{Effect of Stagewise Focused Distillation}

In structure-focused Stage~I, we bias distribution matching away from the low-noise regime using non-uniform timestep sampling. Four schemes (\cref{fig:ablation_prob_curves}) are evaluated when training our 2-step model and we report the results in the top of \cref{tab:ablation_sfd}. Compared with uniform sampling and the low-noise–biased curve~d, curves~a--c that downweight low-noise timesteps consistently perform better. Curve~c performs best, as it also moderately reduces sampling near the pure-noise end, enabling more effective distillation.

However, only structure-focused training in stage~I is not good enough as details generation ability remains suboptimal. See the bottom part of \cref{tab:ablation_sfd}, by further enabling the proposed Detail-focused Adversarial Refinement (Stage~II), the student model obtains consistent gains across all quality metrics, indicating that Stage~II effectively complements Stage I by enhancing fine details.

\subsubsection{Effect of our Cache Design}

We conduct extensive experiments on proposed caching design for extremely few-step distilled models, addressing two questions: Compared with training-free caching applied after distillation, can DCT (\cref{sec:co_training}) recover the quality degradation? Whether the lightweight MLP is necessary for reuse-error compensation?
\begin{figure}[h]
\centering
\definecolor{headviolet}{RGB}{252,228,236}
\definecolor{headpink}{RGB}{222,233,249}
\definecolor{redhi}{RGB}{200,0,0}
\begin{minipage}[t]{0.48\columnwidth}
    \vspace{0pt}
    \centering
    \captionsetup{type=table}
    \caption{Ablation results on cache settings and training variants.}
    \label{tab:ablation_cache}
    \renewcommand{\arraystretch}{1.4}
    \resizebox{\linewidth}{!}{
        \begin{tabular}{ccc |c|cccc}
        \toprule
        \multicolumn{1}{c}{\textbf{Cache}} &
        \multicolumn{1}{c}{\textbf{+Train}} &
        \multicolumn{1}{c}{\textbf{+MLP}} &
        \multicolumn{1}{c}{\cellcolor{headviolet}\textbf{\#NFE}} &
        \multicolumn{1}{c}{\cellcolor{headpink}CLIP$\uparrow$} &
        \multicolumn{1}{c}{\cellcolor{headpink}PS$\uparrow$} &
        \multicolumn{1}{c}{\cellcolor{headpink}IR$\uparrow$} &
        \multicolumn{1}{c}{\cellcolor{headpink}HPS$\uparrow$} \\
        \midrule
        \multicolumn{3}{l|}{SFD w/o cache} & 2.00 & 0.3184 & 22.6995 & 1.1601 & 33.0293 \\
        \midrule
        6 blocks & \xmark & \xmark & 1.75 & 0.3170 & 22.2042 & 1.0152 & 30.9216 \\
        6 blocks & \checkmark & \xmark & 1.75 & 0.3205 & 22.3113 & 1.0500 & 31.0544 \\
        \rowcolor{gray!15}
        6 blocks & \color{redhi}\checkmark & \color{redhi}\checkmark & 1.75 & \textbf{0.3208} & \textbf{22.5161} & \textbf{1.1312} & \textbf{32.2550} \\
        \midrule
        8 blocks & \xmark & \xmark & 1.67 & 0.3183 & 21.7768 & 0.8944 & 29.6062 \\
        8 blocks & \checkmark & \xmark & 1.67 & 0.3198 & 21.9289 & 0.9238 & 30.0994 \\
        \rowcolor{gray!15}
        8 blocks & \color{redhi}\checkmark & \color{redhi}\checkmark & 1.67 & \textbf{0.3204} & \textbf{22.3342} & \textbf{1.0673} & \textbf{31.6850} \\
        \bottomrule
        \end{tabular}
    }
\end{minipage}
\hfill
\begin{minipage}[t]{0.48\columnwidth}
    \vspace{0pt}
    \centering
    \includegraphics[width=\linewidth]{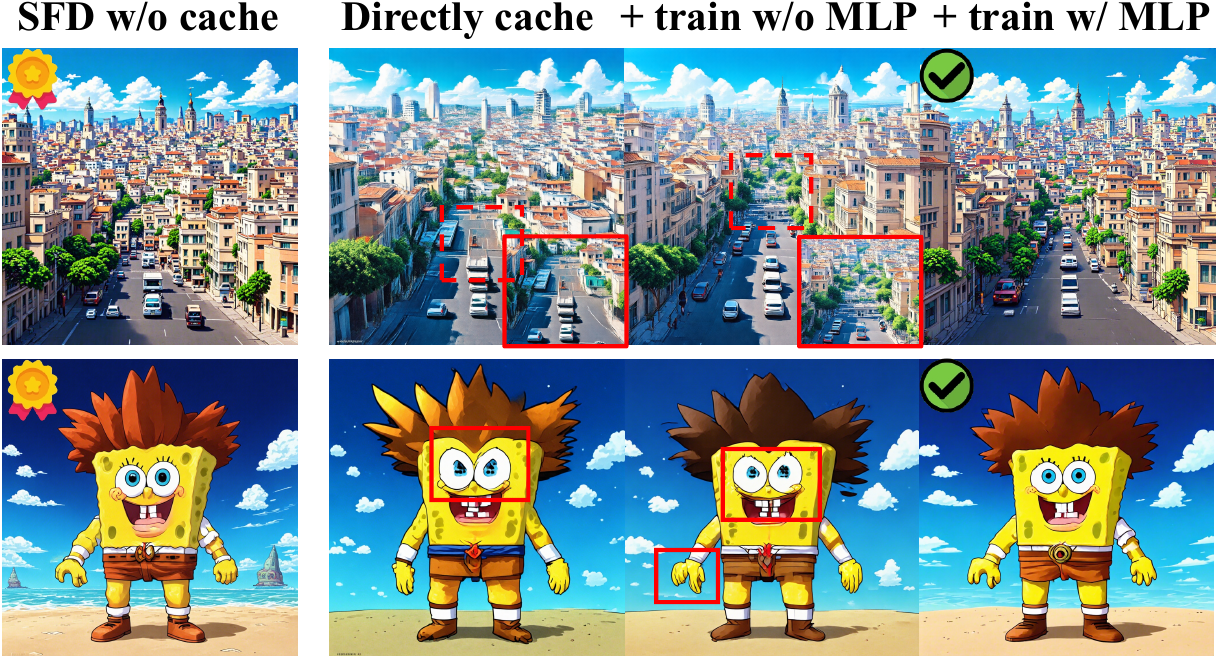}
    \captionsetup{type=figure}
    \vspace{-0.3cm}
    \caption{Qualitative ablation of block-level caching for our DCT.}
    \label{fig:ablation_cache}
\end{minipage}
\end{figure}

As shown in \cref{tab:ablation_cache,fig:ablation_cache}, directly applying in block caching after distillation causes severe quality degradation in both quantitative metrics and visual fidelity, showing that cache acceleration is not an effective plug-and-play component in distilled few-step models. Instead, introducing caching during distillation and optimizing it with DCT largely restores image quality. We further find that if remove the MLP, DCT only partially compensates reuse errors and yields limited recovery. In contrast, by explicitly predicting the reused block contribution, the lightweight MLP significantly improves fidelity, bringing the cached model much closer to the full-computation baseline. These results validate the effectiveness of DCT and the necessity of explicit reuse-error compensation. 

\textbf{Note.} Additional experimental analyses are provided in the supplementary material.

\section{Conclusion}
In this work, we present \textbf{\textit{1.x-Distill}}, the first framework that pushes distribution matching distillation beyond the conventional integer-step regime. To address diversity degradation problem in DMD, we revisit the overlooked role of teacher CFG and introduce a guidance control strategy. We then propose SFD, which decouples structure and detail learning to improve generation quality under extreme step compression. Furthermore, we combine learnable block-level caching into our distillation via DCT.
On SD3-Medium and SD3.5-Large, \textit{1.x-Distill} achieves remarkable performance in both sampling diversity and image quality at 1.67 and 1.74 effective NFE, respectively.

\noindent\textbf{Limitations \& future work.}
While our method demonstrates promising results, its effectiveness on recent larger-scale generative models, such as Qwen-Image(20B)\cite{qwen}, remains to be further explored. In addition, extending \textit{1.x-Distill} to video generation is also an important direction for future work.

%
%
\bibliographystyle{splncs04}
\bibliography{main}

\clearpage
\appendix
\phantomsection

\renewcommand{\thesection}{\Alph{section}}
\renewcommand{\thesubsection}{\thesection.\arabic{subsection}}

\renewcommand{\theHsection}{appendix.\Alph{section}}
\renewcommand{\theHsubsection}{appendix.\Alph{section}.\arabic{subsection}}

{\LARGE\bfseries Appendix\par}

\vspace{1.6em}
\begin{center}
    {\large\bfseries Table of Contents\par}
    \vspace{0.3em}
\end{center}

\setcounter{tocdepth}{2}

\makeatletter
\providecommand{\authcount}[1]{}
\let\l@title\@gobbletwo
\let\l@author\@gobbletwo
\let\l@date\@gobbletwo
\@starttoc{toc}
\makeatother

\clearpage

\hypertarget{target:alg}{}
\section{Algorithm} \label{sec:algorithm}
\cref{alg:train} presents the training pseudocode of our \textit{1.x-Distill}: Stage~I performs structure-focused distribution matching with CFG-controlled teacher guidance, while Stage~II refines fine-grained details via pixel-space adversarial supervision under the cached inference path. 
\cref{alg:inference} presents the inference procedure of \textit{1.x-Distill}.

\begin{algorithm}[H]
\caption{1.x-Distill Training Procedure}
\label{alg:train}
\definecolor{commentcolor}{RGB}{110,154,155}
\newcommand{\pycomment}[1]{\footnotesize\ttfamily\textcolor{commentcolor}{\# #1}}
\makeatletter
\algnewcommand{\LineComment}[1]{%
  \Statex \hspace{\ALG@thistlm}{\footnotesize\ttfamily\textcolor{commentcolor}{\# #1}}%
}
\makeatother 
\begin{algorithmic}[1]
\Require{Pretrained teacher model $\mu_\text{real}$, 2-step generator schedule $S=\{t_0, t_1\}$ (\emph{e.g.}$\{1.0, 0.75\}$), pixel-space discriminator $D$, VAE decoder $V$}
\Ensure{Optimized student generator $G_\theta$ with attached MLP module $f_\phi$}
\State $G_\theta \gets \text{CopyWeights}(\mu_{\text{real}})$ \Comment{{\color{blue}{Initialize generator}}}
\State $\mu_{\text{fake}} \gets \text{CopyWeights}(\mu_{\text{real}})$ \Comment{{\color{blue}{Initialize fake score estimator}}}

\Statex \pycomment{--- Stage I: Structure-focused Distribution Matching ---}
\For{iter $= 1$ to max\_iters\_stage1}
    \State $x_{t_0} \sim \mathcal{N}(0, I)$
    \State Sample $t_i$ from $S$ 
    \State $x_{t_i} \gets \text{BackwardSimulation}(G_\theta, x_{t_0}, t_0 \to t_i)$
    \Comment{{\color{blue}{Get noisy input as DMD2}}}
    \State $\hat{x}_0 \gets G_\theta(x_{t_i})$
    \If{iter mod $TTUR_1 == 0$}
        \LineComment{Update generator $G_\theta$}
        \State $t \sim \text{ImportanceSampling}(0, 1)$ {\color{blue}{\Comment{Importance Sampling \cref{sec:stage1}}}}
        \State $x_t \gets \text{AddNoise}(\hat{x}_0, t)$
        \State $\nabla_{\theta}\mathcal{L}_{\text{DMD}} \gets \text{GradDMD}(\text{CFG controlled}\ {\mu_{\text{real}}}, \mu_{\text{fake}}, x_t)$ \Comment{{\color{blue}{\cref{eq:dmd_grad}, \cref{eq:cfg_ccd}}}}
        \State $G_\theta \gets \text{update}(G_\theta, \nabla_{\theta}\mathcal{L}_{\text{DMD}})$
    \EndIf
    \State $t \sim \mathcal{U}(0, 1)$
    \State $x_t \gets \text{AddNoise}(\text{detach}(\hat x_0), t)$
    \State $\mathcal{L}_{\text{diffusion}} \gets \text{DiffusionLoss}(\mu_{\text{fake}}(x_t, t), \text{detach}(\hat x_0))$
    \State $\mu_{\text{fake}} \gets \text{update}(\mu_{\text{fake}}, \mathcal{L}_{\text{diffusion}})$
\EndFor
\Statex \pycomment{--- Stage II: Detail-focused Adversarial Refinement ---}
\For{iter $= 1$ to max\_iters\_stage2}
    \State $x_{t_0} \sim \mathcal{N}(0, I)$
    \State $x_{t_1}, \Delta_0 \gets G_\theta(x_{t_0},t_0)$
    \State $\hat{x}_0^{\mathrm{cache}} \gets G_\theta(\text{detach}(x_{t_1}),t_1,f_\phi(\Delta_0))$
    \Comment{{\color{blue}{Get $\hat{x}_0^{\mathrm{cache}}$ along inference path}}}
    \If{iter mod $TTUR_2 == 0$} 
        \LineComment{Update generator $G_\theta$, MLP $f_\phi$}
        \State $\mathcal{L}_{\mathrm{Adv}}^{G} \gets -\log D\left(V(\hat{x}_0^{\mathrm{cache}})\right)$
        \State $G_\theta \gets \text{update}(G_\theta, \mathcal{L}_{\mathrm{Adv}}^{G})$
        \State $f_\phi \gets \text{update}(f_\phi, \mathcal{L}_{\mathrm{Adv}}^{G})$
    \EndIf
    \State $\mathcal{L}_{\mathrm{Adv}}^{D} \gets -\log D\left(V(x^{\star}_0)\right)+\log D\left(V(\hat{x}_0^{\mathrm{cache}})\right)$
    \State $D \gets \text{update}(D, \mathcal{L}_{\mathrm{Adv}}^{G})$
\EndFor
\end{algorithmic}
\end{algorithm}
\begin{algorithm}[H]
\caption{1.x-Distill Inference Procedure}
\label{alg:inference}
\definecolor{commentcolor}{RGB}{110,154,155}
\newcommand{\pycomment}[1]{\footnotesize\ttfamily\textcolor{commentcolor}{\# #1}}
\begin{algorithmic}[1]
\Require{Distilled 2-step generator $G_\theta$ with schedule $S=\{t_0, t_1\}$, trained lightweight module $f_\phi$, VAE decoder $V$}
\Ensure{Clean image sample $x_0$}
    \State $x_{t_0} \sim \mathcal{N}(0, I)$
    \State $x_{t_1}, \Delta_0 \gets G_\theta(x_{t_0},t_0)$
    \State $x_0 \gets G_\theta(x_{t_1},t_1,f_\phi(\Delta_0))$
    \State $x_0 \gets V(x_0)$ {\color{blue}\Comment{Decode latents to pixel-space image}}
\end{algorithmic}
\end{algorithm}

\hypertarget{target:impl}{}
\section{Implementation Details} \label{sec:implementation}
\hypertarget{target:disc}{}
\subsection{Discriminator Design} \label{sec:disc_design}
Our discriminator architecture follows the design in~\cite{hypir}, consisting of a frozen ConvNeXt backbone and a lightweight trainable head.
Specifically, we use the ConvNeXt-XXL visual encoder from a pretrained OpenCLIP model\footnote{\texttt{laion/CLIP-convnext\_xxlarge-laion2B-s34B-b82K-augreg-soup}} as the feature extractor, which outputs multi-level feature maps with channel dimensions $\{384, 768, 1536\}$ together with a pooled global feature of dimension $1024$.
On top of these representations, we attach a multi-level discriminator head.
Each intermediate feature map is processed by spectrally normalized convolution layers with LeakyReLU activations and BlurPool downsampling to produce realism predictions at different spatial scales.
The pooled feature is further passed through a linear classifier to obtain a global realism score.
Predictions from all levels are first averaged within each level and then summed across levels to produce the final adversarial signal as \cref{eq:gan}.

\hypertarget{target:train}{}
\subsection{Training Details} \label{sec:train_details}
We implement our \textit{1.x-Distill} framework in PyTorch and train on 8 NVIDIA A100 GPUs. We adopt Fully Sharded Data Parallel (FSDP) to scale training across GPUs and enable mixed-precision training in \texttt{torch.bfloat16} for both efficiency and stability.

Follow the $\text{shift}(3.0)$ from the scheduler of teacher model, we set the generator timestep schedule to $S=\{1.0,\ 0.9,\ 0.75,\ 0.5\}$ for 4-step SFD, and $S=\{1.0,\ 0.75\}$ for 2-step \textit{1.x-Distill}.
For optimization, we employ the \texttt{AdamW} optimizer across all trainable components, including the student generator $G_\theta$, the fake score estimator $\mu_{\text{fake}}$, the pixel-space discriminator $D$, and the  MLP module $f_\phi$. We set the weight decay to $1e-4$ and the exponential moving average coefficients $(\beta_1, \beta_2) = (0, 0.999)$ in Stage~I, $(0.9, 0.95)$ in Stage~II. 
The learning rate and other configurations are listed in \cref{tab:training_config}.
Notably, the total training cost of \textit{1.x-Distill} is about 71 GPU-hours on SD3-Medium (2B) and 104 GPU-hours on SD3.5-Large (8B). By contrast, DMD2 trains for $64 \times 60$ GPU hours on SDXL (3.5B), indicating that \textit{1.x-Distill} is significantly more training-efficient.
\definecolor{stagegray}{gray}{0.93}
\begin{table}[h]
\centering
\caption{\textbf{Training configurations} for SD3-Medium and SD3.5-Large.}
\label{tab:training_config}
\renewcommand{\arraystretch}{1.15}
\begin{tabular*}{0.9\columnwidth}{@{\extracolsep{\fill}}l|cc}
\toprule
\textbf{Config} & \textbf{SD3-Medium} & \textbf{SD3.5-Large} \\
\midrule
Resolution & \multicolumn{2}{c}{$1024\times1024$}  \\
Total Batch Size  & \multicolumn{2}{c}{32} \\
Precision & \multicolumn{2}{c}{\texttt{bfloat16}($t$ in \texttt{float32})} \\
\midrule
\rowcolor{stagegray}
\multicolumn{3}{c}{\textbf{Stage I}} \\
Generator Learning Rate & 5e-6 & 2e-6 \\
Fake Learning Rate & 4e-5 & 2e-5 \\
Guidance Scale $w$ & 7.0 & 7.0 \\
Control Threshold $\alpha$ & 0.94 & 0.94 \\
Generator Update Frequency & 1 & 1 \\
Max Iteration & 1000 & 600 \\
Training Cost & 27 GPU hours & 40 GPU hours \\

\arrayrulecolor{gray!60}\midrule\arrayrulecolor{black}
\rowcolor{stagegray}
\multicolumn{3}{c}{\textbf{Stage II}} \\
Cache config (skip blocks) & [3\textasciitilde8]/[3\textasciitilde10] & [3\textasciitilde12] \\
Generator Learning Rate & 1e-6 & 5e-7 \\
MLP Learning Rate & 1e-3 & 1e-3 \\
Discriminator Learning Rate & 1e-5 & 1e-5 \\
Generator Update Frequency & 3 & 3 \\
MLP Warmup Iteration & 500 & 500 \\
Max Iteration & 2000 & 2000 \\
Training Cost & 44 GPU hours & 64 GPU hours \\
\bottomrule
\end{tabular*}
\end{table}

For adversarial training in Stage~II, the reference images $x^\star$ are generated using an 8-step model distilled for less than 30 GPU hours by our distribution matching method. 
Compared to directly using the teacher model, it reduces the cost of generating $x^\star$ during training to only about 14\% of the original computation. Moreover, the generated images exhibit richer details than those produced by the teacher model, which further improves the effectiveness of adversarial training.

\hypertarget{target:eval}{}
\subsection{Evaluation Details} \label{sec:eval_details}
In this section, we provide additional details on the evaluation metrics and baseline methods to further demonstrate the comprehensiveness and fairness of our comparison.

\subsubsection{Metrics}
We employ a diverse set of evaluation metrics covering distribution fidelity, prompt alignment, perceptual quality, and human preference:
\begin{itemize}
\item \textbf{FID}~\cite{fid} measures the distribution distance between 2 set of images in the Inception feature space. We compute FID between teacher samples and student samples to evaluate generative fidelity after distillation.
\item \textbf{CLIP Score}~\cite{clipscore}. We compute CLIP Score using the CLIP ViT-B/32 model to measure the semantic similarity between generated images and their corresponding text prompts.
\item \textbf{PickScore}~\cite{pickscore}. A learned preference model trained on large-scale human pairwise comparisons, designed to approximate human judgments of overall image quality and prompt consistency.
\item \textbf{Aesthetic Score}~\cite{laion5b}. An aesthetic predictor trained on LAION aesthetic annotations, focusing on visual appeal and photographic quality.
\item \textbf{HPSv2}~\cite{hpsv2} uses a reward model to capture general human preferences for text-to-image generation.
\item \textbf{ImageReward}~\cite{ir} uses a reward model trained with RLHF to jointly evaluates image quality and prompt alignment.
\end{itemize}
Together with DPG-Bench\cite{dpgbench}, LPIPS-based diversity, and user study, we provide a comprehensive evaluation of generation performance from multiple perspectives.

\subsubsection{Baselines}
We compare \textit{1.x-Distill} against a broad set of publicly available few-step diffusion models based on \emph{SD3-Medium} and \emph{SD3.5-Large}.
The evaluated baselines include trajectory-based, distribution-based and combined distillation approaches:
\begin{itemize}
\item \textbf{Hyper-SD}~\cite{hypersd} is a trajectory distillation method that combines consistency trajectory distillation with human feedback learning. 
The released checkpoint of Hyper-SD3-Medium is a LoRA weight that preserves the CFG mechanism. 
In our evaluation, we set the LoRA scale and guidance scale to the default values of $0.125$ and $5.0$, respectively.
\item \textbf{PCM}~\cite{pcm} is a consistency distillation method. 
In our evaluation, we use the official 4-step and 2-step deterministic checkpoints with t $\texttt{shift}=1$.
\item \textbf{Flash}~\cite{flashdiff} is a distillation method that combines distribution matching and adversarial training. 
\item \textbf{LADD (Turbo)}~\cite{ladd} is a latent-space adversarial distillation method applied to SD3.5-Large.
\item \textbf{TDM}~\cite{tdm}is a representative distribution matching distillation method that outperforms DMD2 in quality and efficiency. So we choose it as our main baseline. Since TDM only releases the 4-step distilled model on SD3-Medium, we follow its official code and try our best to distill the 2-step model on SD3-Medium and SD3-Large.
\end{itemize}

\hypertarget{target:exp}{}
\section{Extended Experiments} \label{sec:extended_exp}

\hypertarget{target:comp}{}
\subsection{Compensation Module} 
\label{sec:comp_module}
We further study the learnable error-compensation module \(f(\cdot)\), which is the key component for stabilizing block reuse in our cached few-step inference. 

\subsubsection{Setup}
Since \emph{SD3-Medium} contains 24 DiT blocks, we fix the same cache setting as \emph{1.x-Distill}-slow, where blocks 3--8 in the second denoising step are skipped and approximated. This corresponds to an effective NFE of 1.75.
We study the compensation module from two aspects: different training settings for block-level caching, and different implementations of the compensation module \(f(\cdot)\).

\paragraph{Training settings.}
We first compare three training settings around block-level caching.
\begin{itemize}
\item{Full-computation baseline.}
This variant uses Stage~I and Stage~II with pixel-space adversarial refinement, but without caching. It serves as the reference without 1.x acceleration.

\item Direct cache after distillation. This variant applies block reuse after distillation, without cache-aware adversarial refinement in Stage~II. It evaluates whether caching can be directly applied to the distilled model in a nearly plug-and-play manner.

\item Distill--Cache co-Training (DCT). This is our full training setting, where cache is explicitly incorporated into Stage~II and optimized jointly with the generator along the cached inference path.
\end{itemize}

\paragraph{Compensation module designs.}
Based on the cache-aware training setting above, we further compare several implementations of \(f(\cdot)\).
\begin{itemize}
\item No compensation. The cached contribution is directly reused without learnable correction.

\item Simple residual MLP (segment-level). Our default design, using a lightweight residual MLP with LayerNorm and a two-layer GELU MLP. The hidden dimension is \(2\times\) the input dimension, and the output layer is zero-initialized.

\item Simple residual MLP (per-block). Separate simple MLP predictors for individual block deltas.

\item Deeper residual MLP. A stronger predictor formed by stacking residual MLP blocks (expansion ratio 2, depth 2, dropout 0).

\item Transformer proxy. One native DiT transformer block for residual delta prediction.
\end{itemize}

\subsubsection{Analysis}
The quantitative results are reported in \Cref{tab:suppl_mlp}. We next analyze the effect of cache-aware training and the design choice of the compensation module.

\paragraph{Effect of training strategy.}
\Cref{tab:suppl_mlp} shows that block-level caching is not directly transferable to extremely few-step distilled models. Directly applying cache after distillation reduces NFE and latency, but causes clear degradation on all preference-oriented metrics. This suggests that feature reuse is substantially more difficult in distilled two-step models, where adjacent steps exhibit larger feature drift and direct reuse introduces significant error. In contrast, incorporating cache into Stage~II and optimizing it through DCT largely restores image quality, showing that cache acceleration in this regime must be learned jointly with the generator.

\paragraph{Effect of compensation design.}
The comparison among different compensation modules further shows that explicit learnable correction is necessary for stable cross-step reuse. Even under cache-aware training, directly reusing the cached contribution without \(f(\cdot)\) still leaves a noticeable performance gap, indicating that joint optimization alone is insufficient.

\definecolor{redhi}{RGB}{200,0,0}

\begin{table}[h]
\centering
\caption{\textbf{Ablation of cache-aware training and compensation module designs.} 
\textit{Top}: different training settings for introducing block-level caching. 
\textit{Bottom}: different implementations of the compensation module \(f(\cdot)\) under the full DCT setting.}
\label{tab:suppl_mlp}
\small
\setlength{\tabcolsep}{3.5pt}
\renewcommand{\arraystretch}{1.5}
\resizebox{0.95\columnwidth}{!}{
\begin{tabular}{c|cc|c|cccccc}
\toprule
\multirow{2}{*}{\textbf{Stage I}} 
& \multicolumn{2}{c|}{\textbf{Stage II}} 
& \multirow{2}{*}{\textbf{\(f(\cdot)\)}} 
& \multirow{2}{*}{\textbf{\#NFE}} 
& \multirow{2}{*}{\textbf{Latency(s)$\downarrow$}} 
& \multirow{2}{*}{\textbf{CLIP}$\uparrow$} 
& \multirow{2}{*}{\textbf{PS}$\uparrow$} 
& \multirow{2}{*}{\textbf{IR}$\uparrow$} 
& \multirow{2}{*}{\textbf{HPS}$\uparrow$} \\
& \textbf{Cache} & \textbf{PixGAN} & & & & & & & \\
\midrule
\checkmark & \xmark & \checkmark & -- & 2.00 & 0.7413 & 0.3184 & 22.6995 & 1.1601 & 33.0293 \\
\checkmark & \checkmark & \xmark & -- & 1.75 & 0.6538 & 0.3170 & 22.2042 & 1.0152 & 30.9216 \\
\checkmark & \checkmark & \checkmark & None & 1.75 & 0.6561 & 0.3205 & 22.3113 & 1.0500 & 31.0544 \\
\arrayrulecolor{gray!60}\midrule\arrayrulecolor{black}
\rowcolor{gray!15}
\checkmark & \checkmark & \checkmark & {\color{redhi}\textbf{Simple MLP (seg.)}} & 1.75 & \textbf{0.6617} & \textbf{0.3208} & \textbf{22.5161} & \textbf{1.1312} & \textbf{32.2550} \\
\checkmark & \checkmark & \checkmark & Simple MLP (per-block) & 1.75 & 0.6649 & 0.3196 & 22.4728 & 1.1184 & 32.0713 \\
\checkmark & \checkmark & \checkmark & Deep ResMLP & 1.75 & 0.6685 & 0.3207 & 22.4986 & 1.1249 & 32.1462 \\
\checkmark & \checkmark & \checkmark & 1 DiT block & 1.75 & 0.6768 & 0.3214 & 22.4415 & 1.1197 & 32.3826 \\
\bottomrule
\end{tabular}
}
\end{table}

Among all variants, the \textbf{simple residual MLP (segment-level)} provides the best overall trade-off. It restores most of the lost quality while remaining lightweight and stable to optimize. In comparison, the \textbf{per-block MLP} offers no clear advantage over segment-level prediction, the \textbf{deeper residual MLP} brings only marginal improvement, and the \textbf{Transformer proxy} fails to yield consistent gains despite its higher complexity. These results suggest that the correction needed for cross-step block reuse is relatively simple, and increasing the capacity of \(f(\cdot)\) offers limited practical benefit.

\hypertarget{target:block}{}
\subsection{Block Selection} \label{sec:block_sel}

We further study how to choose cached blocks, since block selection is critical to the quality--efficiency trade-off in our 1.x inference regime.

\subsubsection{Setup}
\begin{figure}[h]
    \centering
    \includegraphics[width=0.8\linewidth]{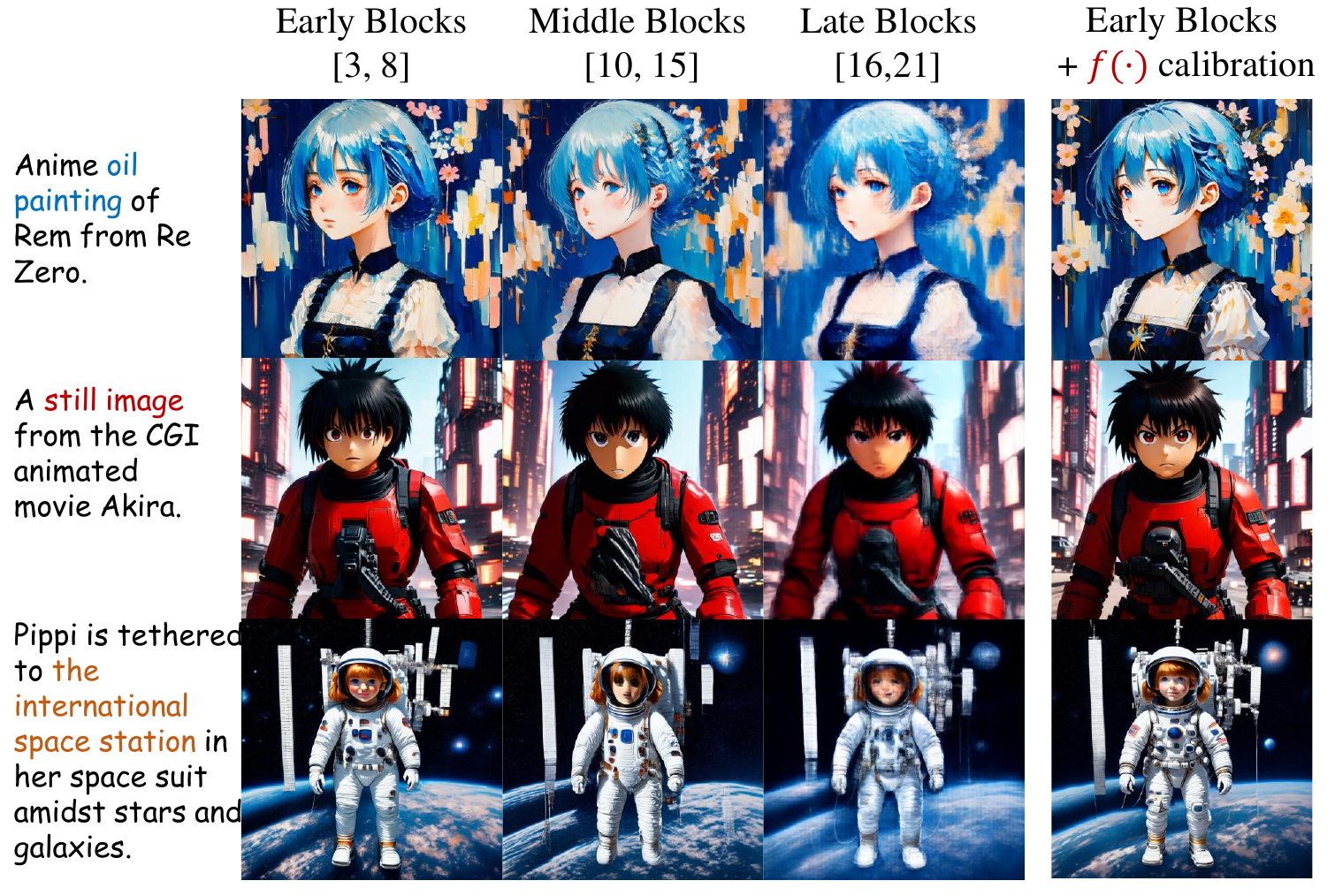}
    \caption{\textbf{Effect of caching different block ranges.} Early-block caching causes relatively mild degradation, while middle and late blocks lead to severe artifacts, consistent with their larger reuse error. With learnable compensation \(f(\cdot)\), caching early blocks largely preserves image quality.}
    \label{fig:suppl_cache}
\end{figure}

All experiments in this section are conducted on \emph{SD3-Medium} distilled to 2-step sampling. We use the simple residual MLP as in the previous subsection Sec.\ref{sec:comp_module} for error compensation and compare five cache settings, including three contiguous ranges, i.e., blocks 3--8, 10--15, and 16--21, and two mixed settings with the same number of cached blocks, i.e., blocks 3--6 with 10--13, and blocks 10--13 with 16--19. To study block sensitivity, we first directly apply cache after Stage~I at inference time, as shown in Fig.\ref{fig:suppl_cache}, and then verify whether the same trend remains after training.

\definecolor{redhi}{RGB}{200,0,0}

\begin{table}[!htbp]
\centering
\caption{\textbf{Ablation of block selection} on \emph{SD3-Medium}.
The first row reports the full two-stage SFD model without cache acceleration. The remaining rows compare different cached block ranges using the same simple residual MLP for error compensation.}
\label{tab:suppl_block}
\small
\setlength{\tabcolsep}{3.5pt}
\renewcommand{\arraystretch}{1.5}
\resizebox{0.9\columnwidth}{!}{
\begin{tabular}{c|c|cccccc}
\toprule
\textbf{Stage} & \textbf{Cached blocks} & \textbf{\#NFE} & \textbf{Latency(s)$\downarrow$} & \textbf{CLIP}$\uparrow$ & \textbf{PS}$\uparrow$ & \textbf{IR}$\uparrow$ & \textbf{HPS}$\uparrow$ \\
\midrule
SFD w/o cache & -- & 2.00 & 0.7413 & 0.3184 & 22.6995 & 1.1601 & 33.0293 \\
\arrayrulecolor{gray!60}\midrule\arrayrulecolor{black}
\rowcolor{gray!15}
DCT & {\color{redhi}\textbf{3--8}} & 1.75 & \textbf{0.6617} & \textbf{0.3208} & \textbf{22.5161} & \textbf{1.1312} & \textbf{32.2550} \\
DCT & 10--15 & 1.75 & 0.6620 & 0.3196 & 22.1887 & 1.0218 & 30.9846 \\
DCT & 16--21 & 1.75 & 0.6621 & 0.3179 & 21.7315 & 0.8734 & 29.2148 \\
DCT & 3--6 + 10--13 & 1.75 & 0.6627 & 0.3201 & 22.3419 & 1.0726 & 31.5327 \\
DCT & 10--13 + 16--19 & 1.75 & 0.6628 & 0.3188 & 21.9642 & 0.9481 & 30.1029 \\
\bottomrule
\end{tabular}
}
\end{table}

\subsubsection{Analysis}
As shown in \ref{fig:method_cache}(a), we measure the block-wise reuse error on \emph{SD3-Medium} as the contribution change across adjacent denoising steps
\[
e_n=\|\Delta_{n,t+1}-\Delta_{n,t}\|_1, \qquad
\Delta_{n,t}=O_{n,t}-I_{n,t}.
\]
Early blocks consistently exhibit smaller reuse error, indicating stronger temporal redundancy, whereas later blocks show much larger reuse error.

The direct-cache results closely follow the reuse-error curve. As shown in \Cref{tab:suppl_block}, caching blocks 3--8 causes the smallest degradation, while caching blocks 10--15 and especially 16--21 lead to much larger quality drop, showing that early blocks are more suitable for reuse than later ones.
The mixed settings show the same trend. Although blocks 3--6 + 10--13 and blocks 10--13 + 16--19 cache the same number of blocks, their performance still differs noticeably, again following the reuse-error curve rather than the cache ratio alone. This suggests that uncached blocks cannot reliably absorb the distortion introduced by high-error cached ranges.

Based on these observations, we choose blocks 3--8 for \emph{1.x-Distill}-slow, and further extend the cached range to blocks 3--10 for the more aggressive \emph{1.x-Distill}-fast setting.

\hypertarget{target:loss}{}
\subsection{Training Objectives for DCT} \label{sec:loss_dct}
We further investigate the effect of incorporating additional knowledge distillation (KD) objectives in Distill--Cache co-Training (DCT). 
Inspired by previous diffusion pruning works, we consider two commonly used KD formulations: a feature-level KD objective and an output-level KD objective.

The feature-level KD objective encourages the predicted block contribution produced by the MLP to match the ground-truth contribution of the skipped blocks:

\begin{equation}
\mathcal{L}_{\text{feat}} 
= 
\mathbb{E}\left\| \Delta_1 - f(\Delta_0) \right\|_2^2 .
\end{equation}

The output-level KD objective directly constrains the prediction of the cached model to match the full-computation model. 
Let $v(x_t,t)$ denote the velocity prediction of the original model and 
$v^{\text{cache}}(x_t,t,f(\Delta_0))$ denote the prediction of the cached model using the reused block contribution predicted by $f(\Delta_0)$. 
The output-level KD loss is defined as:
\begin{equation}
\mathcal{L}_{\text{out}}
=
\mathbb{E}\left\|
v(x_{t_1},t_1) -
v^{\text{cache}}(x_{t_1},t_1,f(\Delta_0))
\right\|_2^2 .
\end{equation}

\begin{center}
\includegraphics[width=0.9\columnwidth]{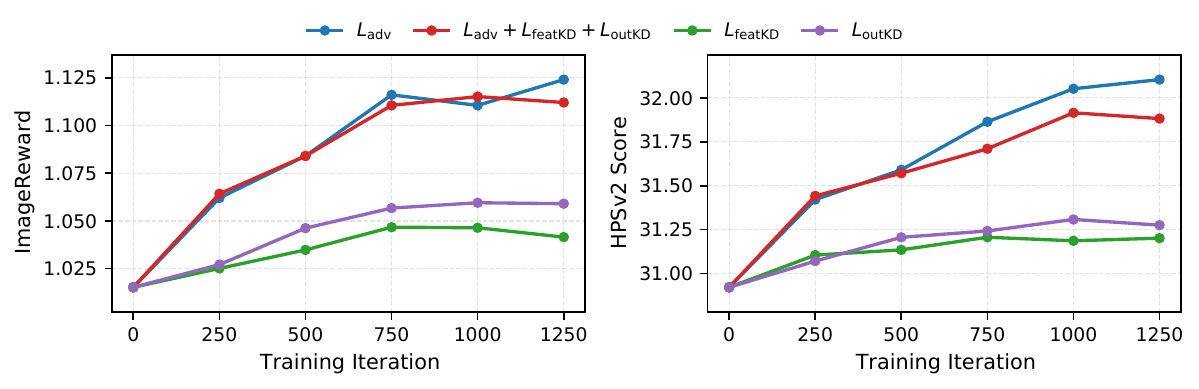}
\captionof{figure}{Effect of different training objectives in DCT. Experiments are conducted on SD3-Medium with a cache block 3--8, using identical training configurations.}
\label{fig:suppl_dct}
\end{center}

We compare different training objectives for DCT, including adversarial loss only ($\mathcal{L}_{\text{adv}}$), adversarial loss with both KD objectives ($\mathcal{L}_{\text{adv}} + \mathcal{L}_{\text{feat}} + \mathcal{L}_{\text{out}}$), and the KD objectives individually. 
As shown in \cref{fig:suppl_dct}, adding feature-level and output-level KD does not provide noticeable improvement, while using the KD objectives alone leads to significantly worse performance. 
These results indicate that pixel-space adversarial supervision already provides an effective signal for correcting cache-induced errors, and additional KD constraints are unnecessary in our setting.

\hypertarget{target:visual}{}
\section{Additional Visual Results} \label{sec:visual}

\begin{figure}[h]
  \centering
    \includegraphics[width=0.9\columnwidth]{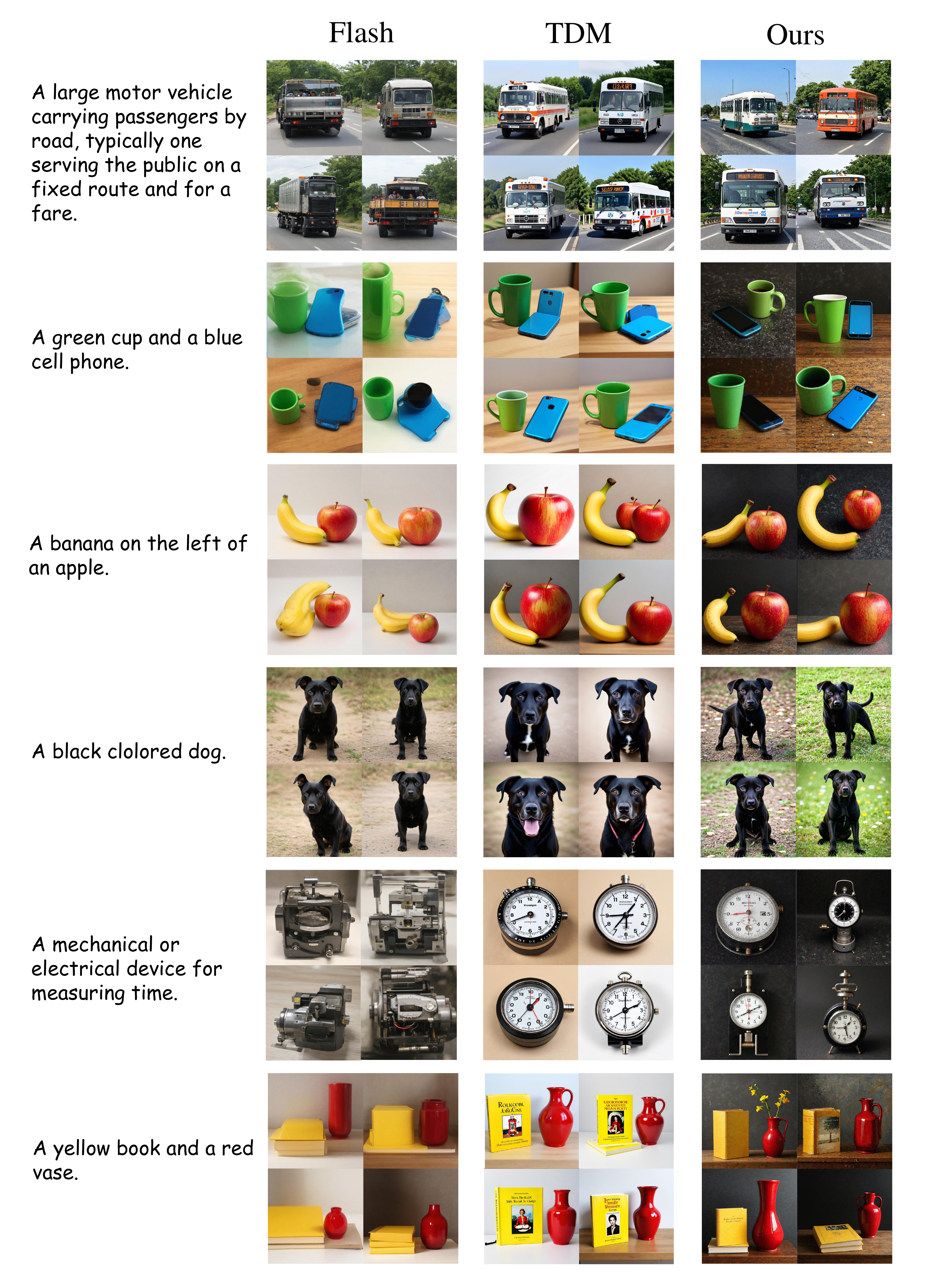} 
  \caption{\textbf{Visual comparison of diversity} under the 4-NFE setting distilled from SD3-Medium. Compared with two distribution-matching baselines, our approach produces more diverse samples while maintaining generation quality and prompt alignment.}
  \label{fig:suppl_diversity}
\end{figure}

\begin{figure}[h]
  \centering
    \includegraphics[width=\columnwidth]{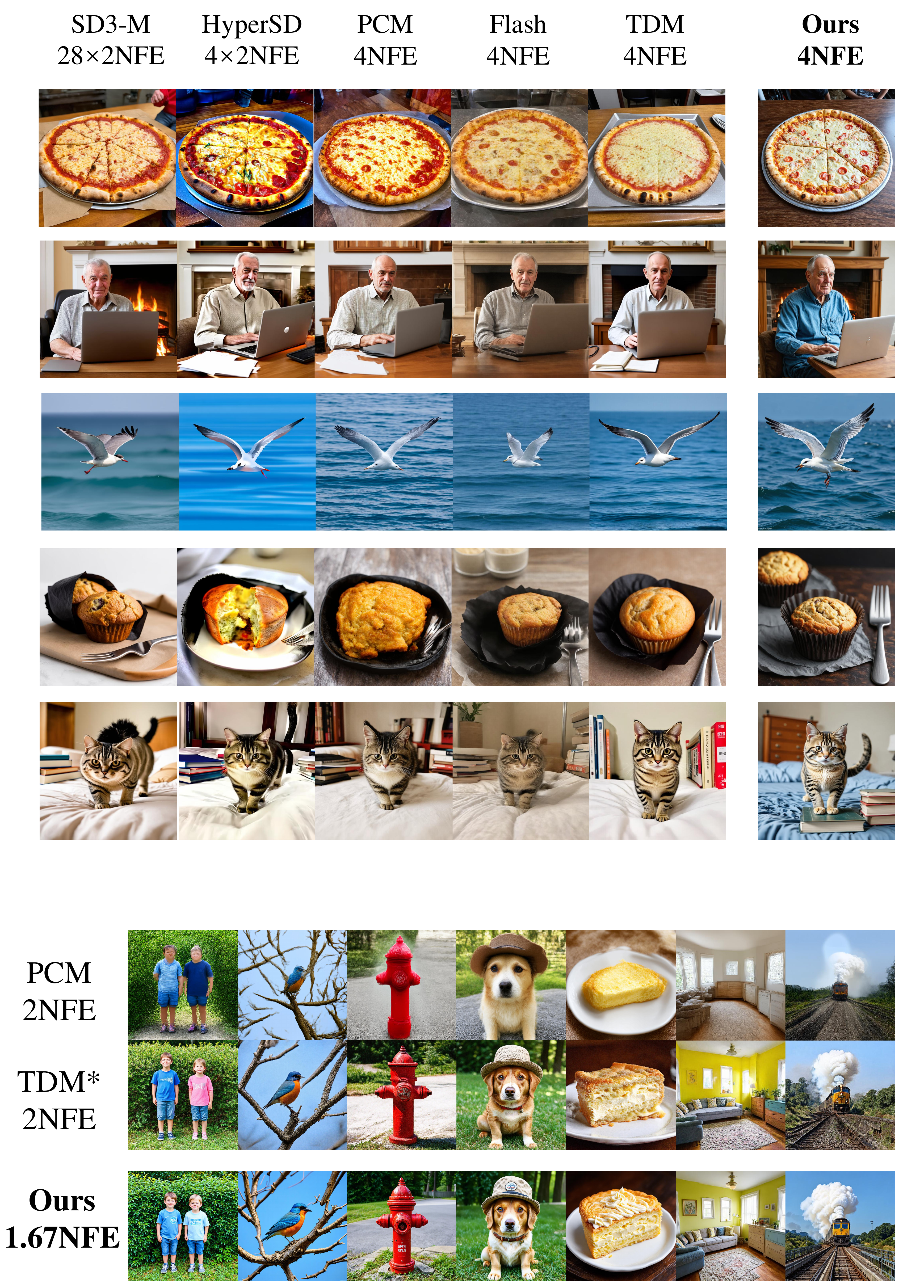} 
  \caption{\textbf{Additional qualitative comparison on SD3-Medium.} Even with only 1.x NFE sampling, \emph{1.x-Distill} produces images with more realistic details than existing few-step baselines. Please zoom in for details.}
  \label{fig:suppl_comp_medium}
\end{figure}

\begin{figure}[h]
  \centering
    \includegraphics[width=\columnwidth]{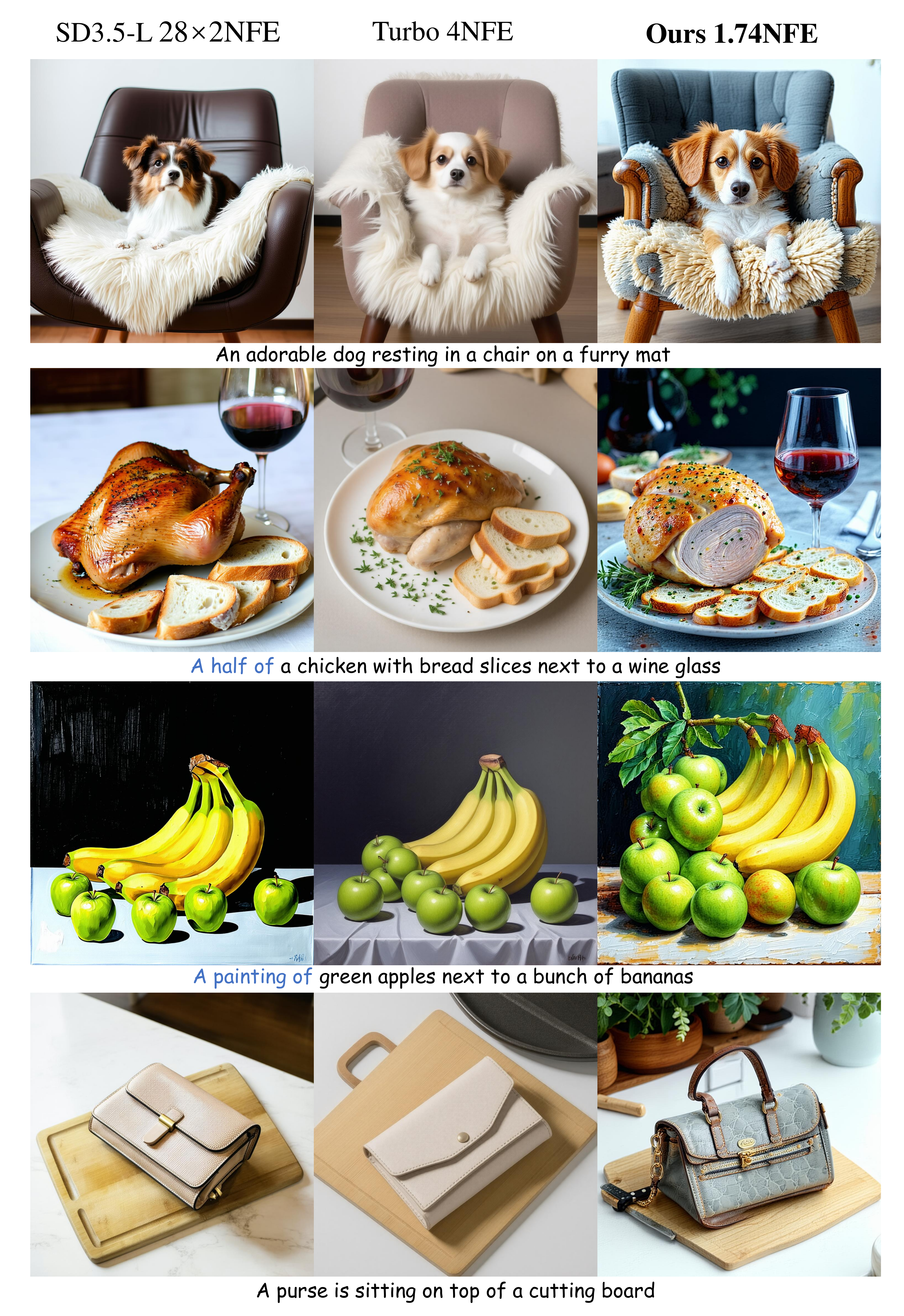} 
  \caption{\textbf{Additional qualitative comparison on SD3.5-Large.} Even with only 1.x NFE sampling, \emph{1.x-Distill} produces images with more realistic details than existing few-step baselines. Please zoom in for details.}
  \label{fig:suppl_comp_large}
\end{figure}

\Cref{fig:suppl_diversity} presents the visual comparison of diversity on SD3-Medium. 
Compared with DMD-like methods, our approach improves sample diversity while maintaining generation quality and prompt alignment. 
Further comparisons on SD3-Medium and SD3.5-Large are provided in \cref{fig:suppl_comp_medium} and \cref{fig:suppl_comp_large}, respectively. 
Even with only 1.x NFE sampling, \emph{1.x-Distill} produces images with rich details and strong visual realism.

\end{document}